\PassOptionsToPackage{table}{xcolor}
\documentclass[acmtog]{acmart}

\usepackage{multirow}
\usepackage{makecell}
\usepackage{microtype}
\usepackage{booktabs}
\usepackage[ruled,vlined]{algorithm2e}

\definecolor{oursgreen}{RGB}{230,245,235}
\definecolor{groupgray}{RGB}{245,246,248}

\SetAlFnt{\small}
\SetAlCapFnt{\small}
\SetAlCapNameFnt{\small}
\SetAlCapHSkip{0pt}

\copyrightyear{2026}
\acmYear{2026}
\setcopyright{cc}
\setcctype{by}
\acmConference[SA Conference Papers '26]{SIGGRAPH Asia 2026 Conference Papers}{December 01--04, 2026}{Kuala Lumpur, Malaysia}
\acmBooktitle{SIGGRAPH Asia 2026 Conference Papers (SA Conference Papers '26), December 01--04, 2026, Kuala Lumpur, Malaysia}
\acmDOI{10.1145/3829340.3842199}
\acmISBN{979-8-4007-2842-6/2026/12}

\begin{document}
\title{SkyAnchor: Updating Metric-scale Aerial 3D Gaussian Scenes from Unposed Ground-View Sequences}

\author{Zhuoxiao Li}
\authornotemark[1]
\affiliation{%
  \institution{Hong Kong University of Science and Thecnology (Guangzhou)}
  \city{Guangzhou}
  \state{Guang Dong}
  \country{China}
}
  
\author{Xinyi Liu}
\authornotemark[1]
\affiliation{%
  \institution{Hong Kong University of Science and Thecnology (Guangzhou)}
  \city{Guangzhou}
  \state{Guang Dong}
  \country{China}
}

\author{Taoyu Wu}
\affiliation{%
  \institution{University of Liverpool}
  \city{Liverpool}
  \country{United Kindom}}


\author{Yinrui Ren}
\affiliation{%
  \institution{Hong Kong University of Science and Thecnology (Guangzhou)}
  \city{Guangzhou}
  \state{Guang Dong}
  \country{China}
}
\author{Tongyan Hua}
\affiliation{%
  \institution{Hong Kong University of Science and Thecnology (Guangzhou)}
  \city{Guangzhou}
  \state{Guang Dong}
  \country{China}
}
\author{OU Jing}
\affiliation{%
  \institution{Hong Kong University of Science and Thecnology (Guangzhou)}
  \city{Guangzhou}
  \state{Guang Dong}
  \country{China}
}
\author{Shuai Zhang}
\affiliation{%
  \institution{Hong Kong University of Science and Thecnology (Guangzhou)}
  \city{Guangzhou}
  \state{Guang Dong}
  \country{China}
}
\author{Dongli Wu}
\affiliation{%
  \institution{Hong Kong University of Science and Thecnology (Guangzhou)}
  \city{Guangzhou}
  \state{Guang Dong}
  \country{China}
}

\author{Rongjun Qin}
\affiliation{%
  \institution{The Ohio State University}
  \city{Ohio}
  \country{United States of America}}

\author{Ge Lin KAN} 
\affiliation{%
  \institution{Hong Kong University of Science and Thecnology (Guangzhou)}
  \city{Guangzhou}
  \state{Guang Dong}
  \country{China}
}
  
\author{Wufan Zhao}\authornotemark[2]
\affiliation{%
  \institution{Hong Kong University of Science and Thecnology (Guangzhou)}
  \city{Guangzhou}
  \state{Guang Dong}
  \country{China}
}

\begin{abstract}

We study how to update a pre-built aerial scene with a newly captured, unposed ground-view sequence. The aerial scene already contains a reliable metric Structure-from-Motion (SfM) reconstruction and a pre-trained 3D Gaussian Splatting (3DGS) model, whereas the ground-view sequence is collected later to add street-level appearance but has unknown camera poses and global scale.
Registering this sequence to the aerial SfM reconstruction is challenging because single-image cross-view localization is brittle and long trajectories are prone to drift. To address these challenges, we present SkyAnchor, which treats the existing aerial scene as a fixed scaffold for ground-view registration and scene update instead of jointly reconstructing aerial and ground imagery from scratch. It first localizes short groups of consecutive ground frames against geometrically verified aerial support, producing sparse anchor poses. It then recovers the full ground trajectory with anchor-constrained submaps, fixing the front and rear anchor poses during incremental registration and bundle adjustment. Finally, it inserts filtered ground Gaussians while preserving the aerial view, followed by lightweight joint refinement. Experiments on seven real aerial--ground scenes show accurate metric ground trajectories  and updated 3D Gaussian scenes with strong aerial- and ground-view rendering quality. Project Page: \href{https://github.com/ZhuoxiaoLi/skyanchor}{{Skyanchor}}. 

\end{abstract}

%
%


%
%


\begin{teaserfigure}
    \centering
  \includegraphics[width=1\textwidth]{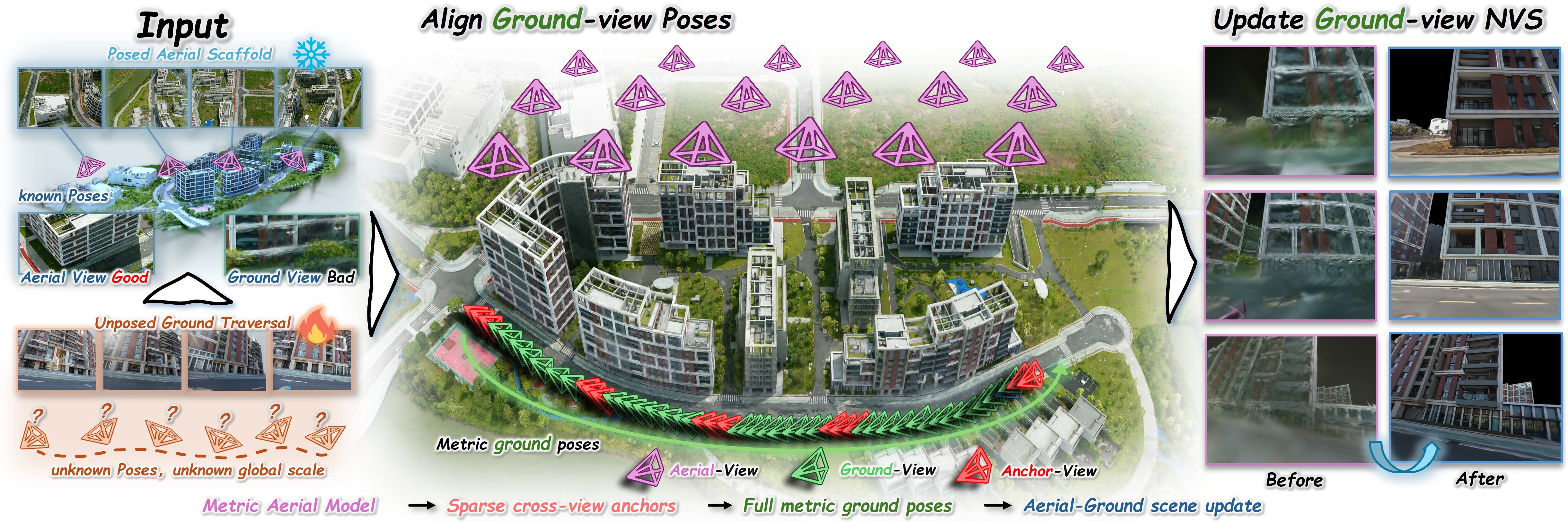}
  \caption{\textit{SkyAnchor} takes a \textbf{posed} aerial scaffold and an \textbf{\underline{unposed}} ground-view sequence as input. By localizing sparse anchor groups and propagating metric constraints to the full ground trajectory, it recovers ground camera poses in the aerial metric frame and updates the scene with street-level appearance.}
  \label{fig:teaser}
\end{teaserfigure}

\maketitle

\section{Introduction}
Large-scale 3D Gaussian Splatting \cite{kerbl2023gaussian} reconstruction increasingly relies on aerial imagery to provide wide-area coverage and {an aerial metric frame} \cite{liu2024citygaussian, vastgaussian, jiang2025horizongs,zhang2025crossviewgs}. Aerial surveys collected by drones or aircraft can efficiently capture building layouts, terrain structure, and overhead appearance \cite{zhou202uavphoto, wang2025uavscenes, YANG2015262isprs1, jensen2014dtu, seitz2006mvs}. Ground-view imagery complements aerial surveys with facade and pedestrian-level details, enabling richer scene models and more realistic rendering \cite{li2023matrixcity,chen2024gigags,nex2015isprsbenchmark,LIAO2024173isprs9}. However, updating a monocular ground-view sequence into {a existing aerial scene} presents substantial challenges: the ground frames are initially unposed, their global scale is unknown, and viewpoint and appearance differences between aerial and street-level imagery are significant \cite{zhu2021vigor,yan2023render2loc,wu2026set}.

This update setting is different from standard joint aerial--ground reconstruction. Existing pipelines typically reconstruct aerial and ground imagery together through cross-view pose estimation, joint SfM, or mixed-view scene optimization \cite{vuong2025aerialmegadepth,zhang2024ucgs,jiang2025horizongs,zhang2025crossviewgs}. They are useful for one-shot offline modeling, but it is poorly matched to map maintenance: each new ground-view sequence would require rerunning a large mixed-view reconstruction, and the optimization may perturb {an aerial scene} that is already the trusted metric reference. {Therefore, {the aerial scene} should not be treated as another unknown input; {it should remain as a fixed scaffold} when new ground capture is registered.}

However, registering an unposed ground-view sequence to such aerial scaffold introduces three coupled technical challenges. First, single-image aerial--ground localization is brittle under large viewpoint, scale, and appearance changes \cite{zhu2021vigor,mi2024congeo,wu2026set}, so a few isolated cross-view matches are not reliable enough to initialize a long sequence. Second, even correct sparse poses do not prevent long-range drift, because metric errors accumulate as trajectory recovery moves away from the localized gorund poses \cite{wu2013linear,schonberger2016sfm}. Third, scene update is asymmetric: the aerial scaffold already represents global structure, while the ground-view sequence supplies local details that should be inserted without degrading the aerial Gaussian model. 

We therefore present SkyAnchor, a pipeline designed for incremental aerial--ground scene updates. Rather than treating {the aerial model} as another unknown input, SkyAnchor {keeps it fixed} while registering new ground observations. First, we partition the ground-view sequence into submaps and localize these anchor groups at the {front and rear} of each submap by exploiting their internal multi-view geometry, thereby establishing sparse {anchor poses in the metric aerial scaffold}. These anchor poses then serve as robust reference that mitigate drift during subsequent trajectory recovery. Next, the full ground trajectory is recovered via anchor-constrained submaps, where {front and rear anchor poses} are kept fixed during incremental registration and bundle adjustment to propagate metric scale throughout the sequence. Finally, the scene is updated by inserting filtered ground geometry and optimizing {ground Gaussians}. A lightweight joint refinement reconciles any residual inconsistencies, producing an updated scene that supports high-quality rendering from both aerial and ground viewpoints.

We evaluate SkyAnchor on seven real {aerial--ground} scenes and the experimental results show that SkyAnchor improves anchor localization robustness, reduces trajectory drift, and produces strong rendering quality from both aerial and ground viewpoints. 

Our contributions are:

\begin{itemize}
    \item We formulate ground-view updating, where an unposed and scale-ambiguous ground-view sequence is registered to {a fixed aerial scaffold} and used to update it without rerunning mixed-view pose reconstruction. 
    
    \item We introduce {an anchor localization stage} that registers short groups of consecutive ground frames to geometrically verified aerial support, improving robustness over single-frame localization.
    
    \item We propose anchor-constrained trajectory recovery, which keeps sparse anchor poses fixed during submap growth and bundle adjustment to propagate aerial metric constraints through long ground-view sequences.
    
    \item We demonstrate through extensive experiments that our scene update strategy produces high-quality 3D Gaussian scenes, preserving the aerial scaffold while enriching street-level appearance.
\end{itemize}

\begin{figure*}[t]
    \centering
    \includegraphics[width=1\linewidth]{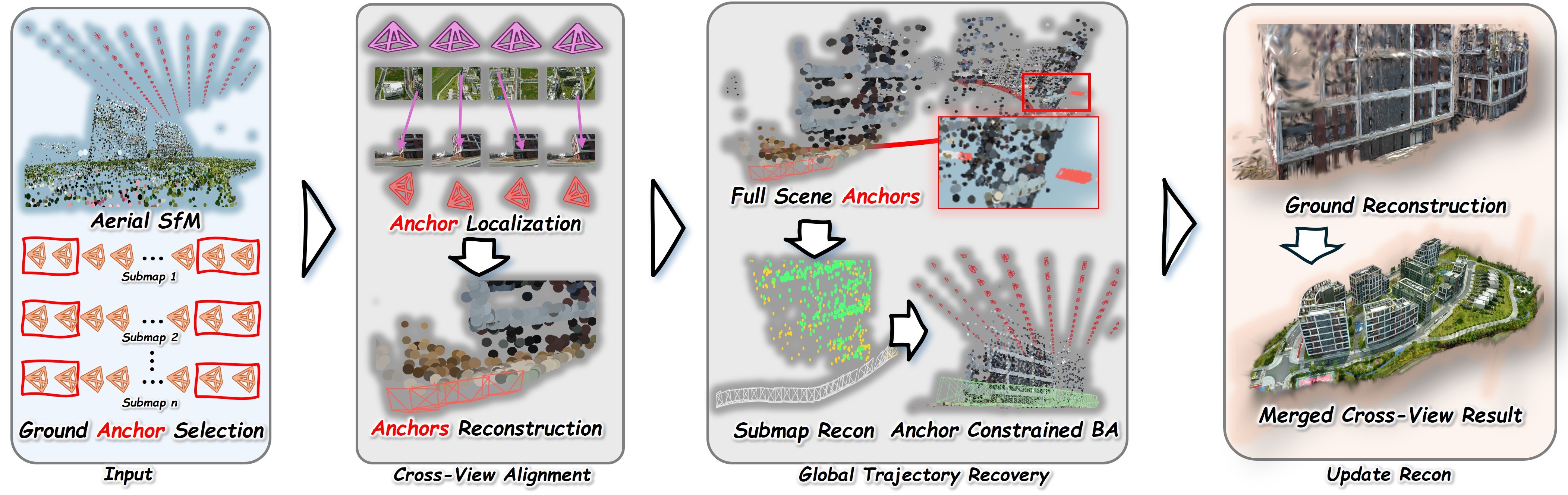}
    \caption{\textbf{Overview of SkyAnchor.} A posed aerial scaffold provides the metric reference. SkyAnchor localizes sparse anchor groups, recovers the full ground trajectory with anchor-constrained submaps, and updates the aerial Gaussian model with street-level geometry and appearance.}
    \label{fig:pipe}
\end{figure*}
\section{Related Work}
\subsection{Cross-View Localization and Registration.}
Cross-view localization estimates the pose or place of ground imagery relative to overhead observations \cite{mi2024congeo,wu2026set}. Retrieval-based benchmarks and methods such as VIGOR \cite{zhu2021vigor}, ConGeo \cite{mi2024congeo}, and Set-CVGL \cite{wu2026set} improve robustness to viewpoint and appearance changes in a shared feature space, while geometry-driven systems such as Render-and-Compare \cite{yan2023render2loc} and Metadata-free Georegistration \cite{bredvik2025metadatafree} refine explicit aerial--ground poses using rendering or synthetic views. These methods are useful for obtaining initial cross-view correspondences, but they are usually defined per image, per query set, or with a pose prior. 

\subsection{Pose-Free and Incremental Reconstruction.}
Pose-free reconstruction systems recover geometry from unordered collections or image streams without calibrated input poses \cite{lan2025stream3r,zhang2026LoGeR,xie2026scal3r,zhuo2025streamvggt}. DUSt3R \cite{wang2024dust3r} and MASt3R \cite{leroy2024mast3r} predict dense pointmaps and matches; MASt3R-SfM \cite{duisterhof2024mast3rsfm} and MP-SfM \cite{pataki2025mpsfm} build scalable SfM pipelines from these priors; and SLAM-based 3R\cite{liu2025slam3r} and Stream-based 3R \cite{li2025streamgs} extend related ideas to sequential reconstruction. These systems are strong baselines for unposed ground imagery, but they usually recover the sequence in a free coordinate system and treat all inputs symmetrically \cite{wang2025cust3r,chen2025ttt3r,murai2025mast3r_slam,deng2025vggt_long,yuan2026infinitevggt}. 

\subsection{Aerial--Ground Reconstruction and View Synthesis.}
In large-scale novel-view synthesis \cite{barron2021mipnerf,meganerf,mi2023switchnerf}, accurate camera poses \cite{liu2024citygaussian, chen2024dogaussiandistributedorientedgaussiansplatting, li2024mvgsplatting} and a consistent scene representation  are essential for rendering photorealistic images from unseen viewpoints \cite{tancik2022blocknerf, barron2021mipnerf, yu2024mipsplatting,cheng2024gaussianpro, seo2024flod} and extract fine geometry \cite{guedon2023sugar,wolf2024surface_3dgs,li2023neuralangelo, huang20242dgs, chen2024pgsr, zhang2024rade, yan2023render2loc}. 
Aerial--ground reconstruction methods jointly estimate cross-view geometry and scene representations from mixed observations. The AerialMegaDepth dataset \cite{vuong2025aerialmegadepth} studies scalable aerial--ground camera estimation and view synthesis; DRAGON \cite{ham2024dragon}, UC-GS \cite{zhang2024ucgs}, Horizon-GS \cite{jiang2025horizongs}, and CrossView-GS \cite{zhang2025crossviewgs} improve Gaussian reconstructions by bridging drone and ground views, using uncertainty-aware training, or designing mixed-view scene models. Their common assumption is that aerial and ground images are optimized together for a one-shot reconstruction. 
\begin{figure}[!t]
    \centering
    \includegraphics[width=1\linewidth]{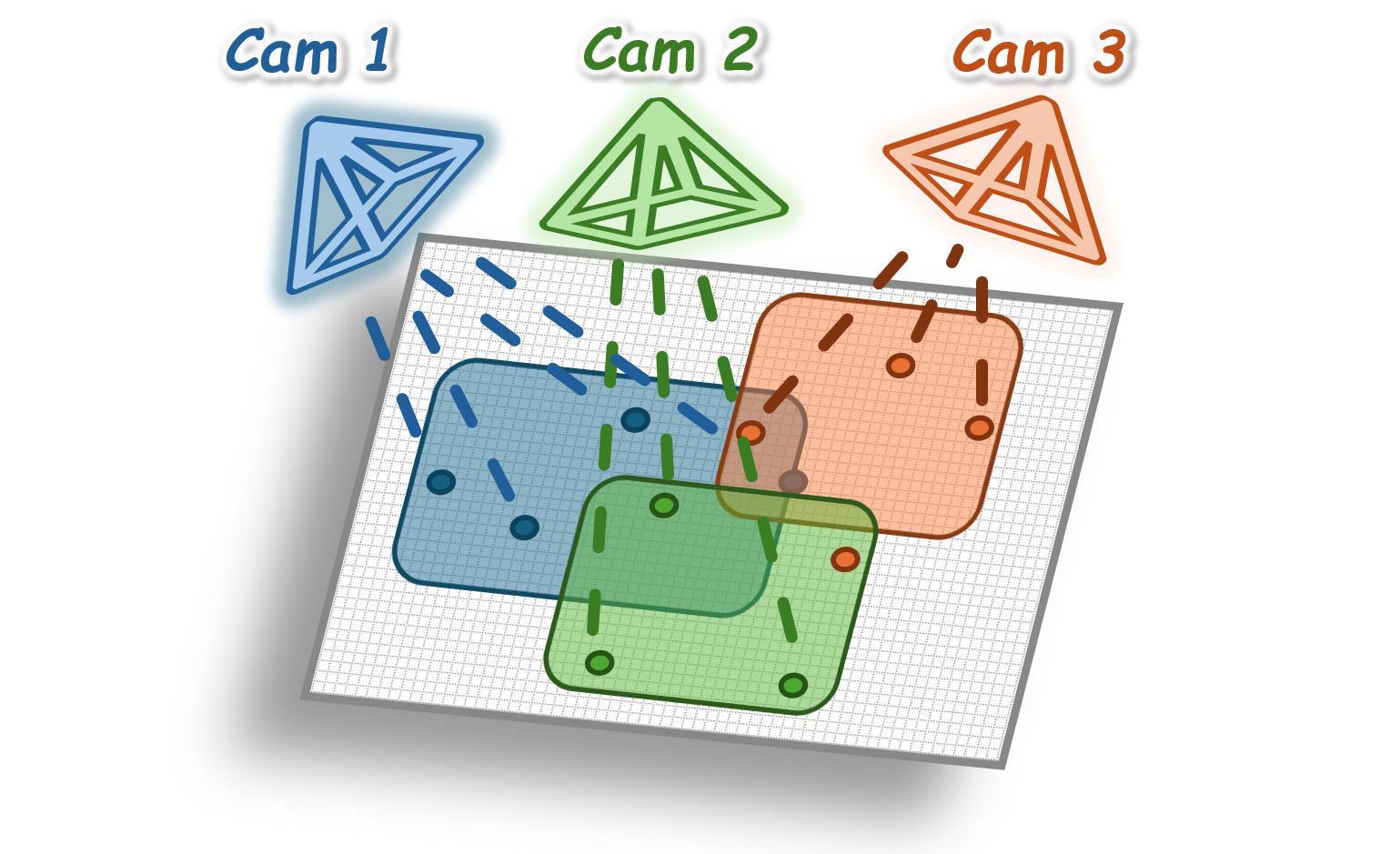}
    \caption{\textbf{Aerial visibility graph construction.} Adjacency in {$G_{\mathrm{vis}}^a$} is determined by the overlap of spatial footprints {$\mathcal{F}_i$} on a common plane.}
    \label{fig:vis_graph}
\end{figure}
\section{Method}
\noindent\textbf{Problem setup.}
{Let $\mathcal{M}^a$ denote the aerial SfM reconstruction comprising aerial images $\mathcal{I}^a=\{I_j^a\}$, reference camera poses $\{\bar{\mathbf{T}}_j^a\}$, and sparse geometry. Its metrically aligned aerial Gaussian model is $\mathcal{G} ^a$. Together, $\mathcal{M}^a$ and $\mathcal{G} ^a$ form the aerial scaffold. Within this scaffold, $\mathcal{M}^a$ remains the fixed geometric and coordinate reference, whereas $\mathcal{G} ^a$ is frozen during ground-content insertion and adjusted only in the final limited joint refinement. Let $\mathcal{I}^g=\{I_1^g,\dots,I_T^g\}$ be a later-collected monocular ground-view sequence with unknown poses $\{\mathbf{T}_t^g\}$ and unknown global scale. Our goal is to recover the ground trajectory in this frame and augment the scaffold with street-level content without rerunning mixed-view pose reconstruction. As shown in Fig.~\ref{fig:pipe}, SkyAnchor first localizes sparse anchor groups, then recovers the trajectory between anchors, and finally recover the ground observation into the aerial Gaussian model.}

\subsection{{Cross-View Anchor Alignment}} \label{sec:anchor}
The first stage partition ground sequences into local submaps and then localizes short anchor (front and end) groups. 

\subsubsection{Aerial Visibility Graph}
{We construct a directed aerial visibility graph so that a retrieved aerial seed can be expanded into a small spatially overlapping neighborhood. For each aerial image $I_i^a$, we project its visible 3D points onto the ground plane and rasterize them into a binary footprint $\mathcal{F}_i$. We retain the $N_{\mathrm{vis}}$ highest-scoring neighbors of each image. We define the footprint IoU score $w_{ij}$ and retained neighbor set as}
\begin{equation}
{
\mathcal{N}_{\mathrm{vis}}(i) = \operatorname{Top}_{N_{\mathrm{vis}}}\!\left(\{j\mid j\neq i\};w_{ij}\right).
}
\end{equation}
{For each $j\in\mathcal{N}_{\mathrm{vis}}(i)$, we add the directed edge $I_i^a\!\rightarrow I_j^a$ with $w_{ij}$ to $G_{\mathrm{vis}}^a$. Although $w_{ij}=w_{ji}$, independently retaining the top $N_{\mathrm{vis}}$ neighbors of each image can produce asymmetric neighborhoods and hence a directed graph. As illustrated in Fig.~\ref{fig:vis_graph}, the weighted graph provides a footprint-overlap proxy for local aerial co-visibility and later supplies nearby support for each selected aerial seed.}

\subsubsection{Anchor Group Selection}
We define an anchor group as $N$ consecutive ground frames, {$A_t=\{I_t^g,I_{t+1}^g,\dots,I_{t+N-1}^g\}$, where $N$ is the anchor-group size.}
We partition the ground-view sequence into submaps and select \textbf{6 front and 6 rear} anchors from the  of each submap. Each group is long enough to provide stable internal geometry but short enough to remain locally rigid.

\subsubsection{Anchor Localization}
{As illustrated in Fig. \ref{fig:AG_recon}, we estimate the metric-scale poses of each anchor through a three-stage pipeline.
\begin{figure}[!t]
    \centering
    \includegraphics[width=1\linewidth]{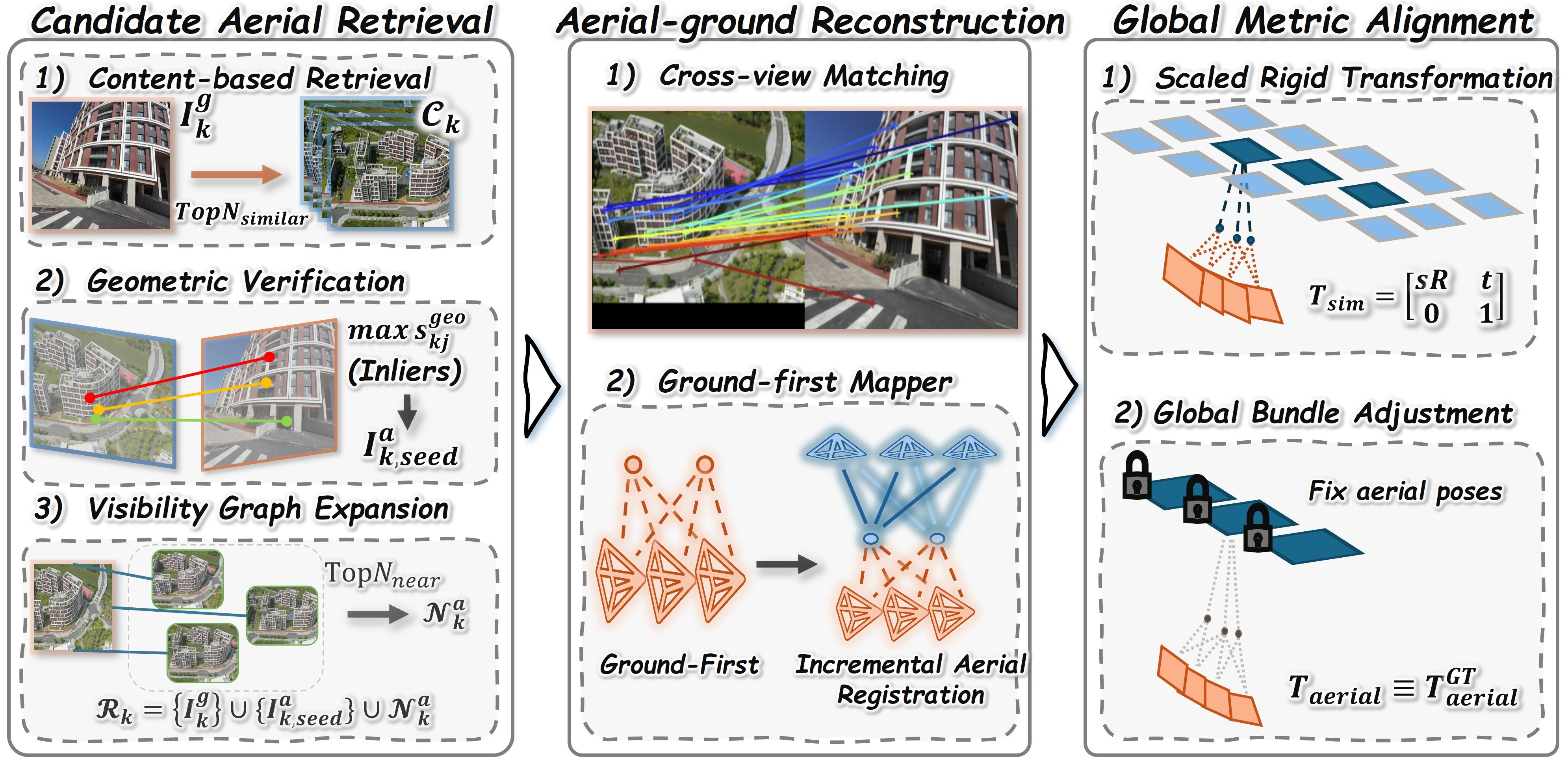}
    \caption{\textbf{Overview of the anchor localization pipeline.} }
    \label{fig:AG_recon}
\end{figure}

\paragraph{Candidate aerial image retrieval.} {For each ground image $I_k^g\in A_t$, we retrieve the $N_{\mathrm{ret}}$ most similar aerial images by descriptor matching and select $I_{k,\mathrm{seed}}^a$ as the candidate with the largest two-view RANSAC inlier count. We write $\operatorname{Nbr}(I)$ for the up-to-$N_{\mathrm{near}}$ outgoing neighbors of aerial image $I$ in $G_{\mathrm{vis}}^a$, ranked by their stored overlap weights. Here $N_{\mathrm{vis}}$ controls the number of neighbors stored for each graph node, whereas $N_{\mathrm{near}}$ controls how many are returned for each seed. The support set for the entire anchor group is}
\begin{equation}
{
\mathcal{R}_t
= \bigcup_{I_k^g\in A_t}
\left(\{I_k^g,I_{k,\mathrm{seed}}^a\}
\cup \operatorname{Nbr}(I_{k,\mathrm{seed}}^a)\right).}
\end{equation}
{The resulting $\mathcal{R}_t$ is the mixed aerial--ground image set passed directly to local reconstruction. Pooling support over the anchor group keeps each aerial neighborhood local while allowing different ground frames to contribute complementary evidence.}

\paragraph{Aerial--Ground Local Reconstruction.} We {run a local aerial--ground reconstruction on $\mathcal{R}_t$} using a modified MP-SfM \cite{pataki2025mpsfm} pipeline with two anchor-specific choices. First, we use the MASt3R checkpoint fine-tuned on AerialMegaDepth \cite{leroy2024mast3r,vuong2025aerialmegadepth} to obtain denser aerial--ground correspondences. Second, we use a {ground-first reconstruction schedule}: recover the internal geometry of the anchor group first, then register the aerial images to this {local reconstruction}. This ordering prevents weak cross-view matches from dominating the early reconstruction. {The result contains local aerial poses $\{\hat{\mathbf{T}}_j^a\}$, local ground poses $\{\hat{\mathbf{T}}_k^g\}$, and sparse 3D points in a common but arbitrarily scaled local frame.}

\paragraph{Global Metric Alignment.} {The preceding local reconstruction has an arbitrary similarity gauge, its ground poses cannot yet serve as metric anchors. We therefore estimate $\mathbf{S}_t\in\mathrm{Sim}(3)$ from corresponding camera centers of the reconstructed aerial poses $\{\hat{\mathbf{T}}_j^a\}$ and reference aerial poses $\{\bar{\mathbf{T}}_j^a\}$, and apply $\mathbf{S}_t$ to every reconstructed camera and 3D point.}
{Starting from this alignment, we run a final bundle adjustment (BA) over the local reconstruction, fixing every reconstructed aerial camera at its reference pose $\bar{\mathbf{T}}_j^a$ and refining the ground cameras and sparse points. We accept the local reconstruction only if it contains enough registered aerial images and satisfies the reprojection-error and similarity-fit-residual thresholds. In Fig. \ref{fig:anchor_reconstruction_vis}, the refined ground poses become the localized anchor poses $\{\mathbf{T}_{i,\mathrm{anc}}^g\}$ used as fixed boundary conditions in trajectory recovery.}
\begin{figure}[!t]
    \centering
    \includegraphics[width=1\linewidth]{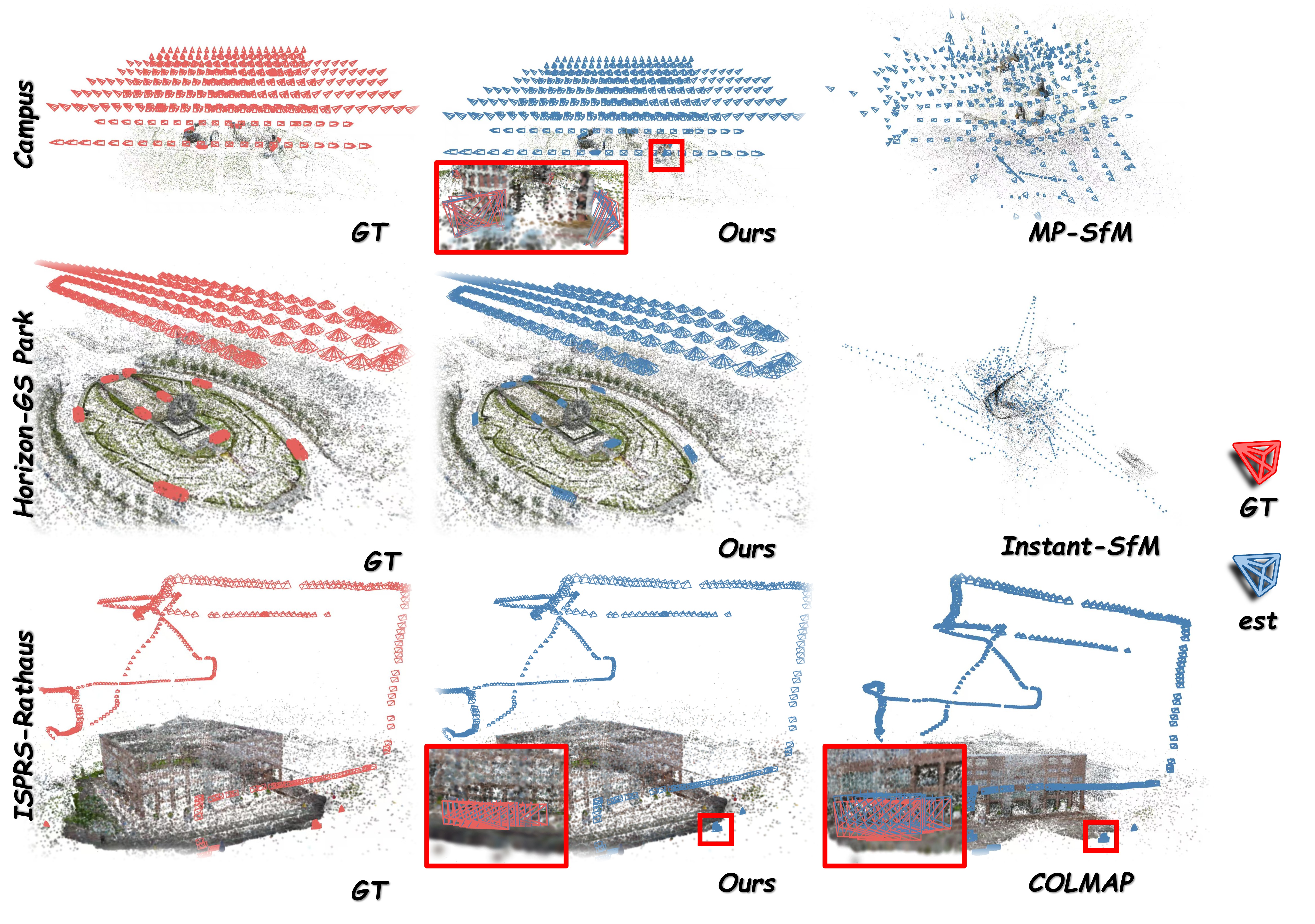}
    \caption{\textbf{{Anchor localization.}} SkyAnchor provide stronger geometric support than other baselines for registering ground observations.}
    \label{fig:anchor_reconstruction_vis}
\end{figure}
\begin{figure}[!t]
    \centering
    \includegraphics[width=1\linewidth]{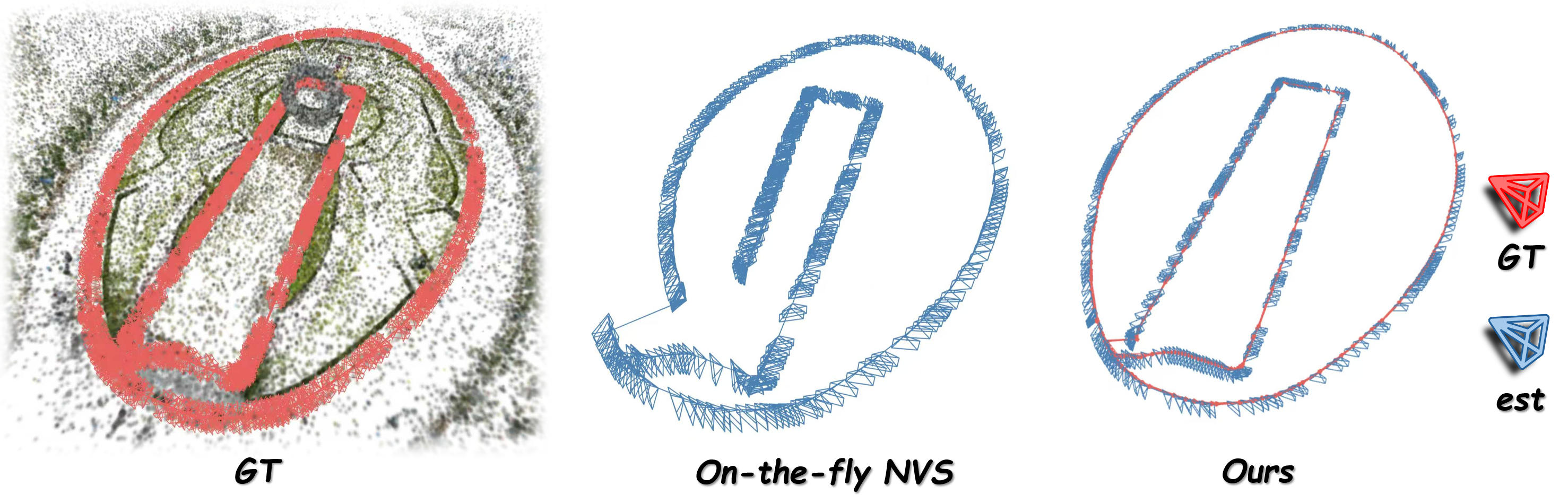}
    \caption{\textbf{{Anchor-constrained trajectory recovery.}} Anchor-constrained submaps reduce drift compared with propagating poses from sparse initialization alone.}
    \label{fig:trajectory_comparison}
\end{figure}
\begin{figure*}[!ht]
    \centering
    \includegraphics[width=1\linewidth]{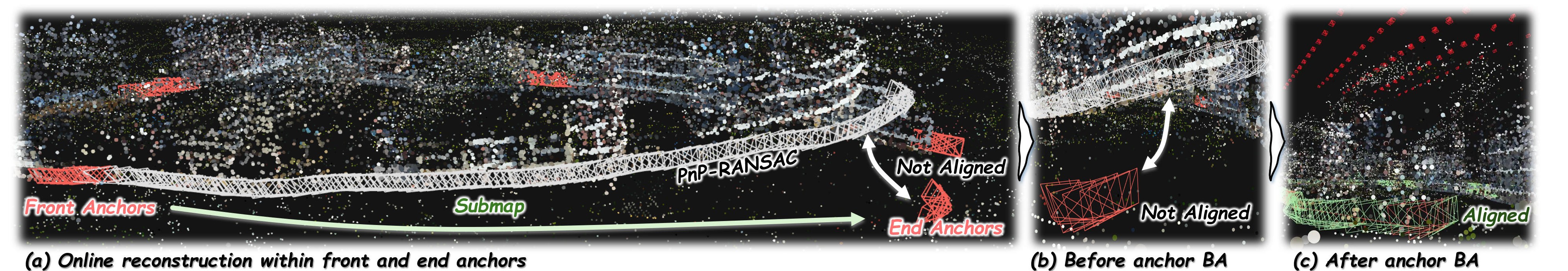}
    \caption{\textbf{{Anchor-constrained Bundle Adjustment.}} A localized submap is aligned to the aerial scaffold and refined while keeping its anchor poses fixed.}
    \label{fig:anchor_ba}
\end{figure*}

\subsection{Anchor-Constrained Trajectory Recovery}

This stage recovers the full ground trajectory by treating {localized anchor poses} as submap boundary conditions (see Fig. \ref{fig:trajectory_comparison}). As shown in Fig.~\ref{fig:anchor_ba}(a), the localized anchor groups split the ground-view
sequence into contiguous submaps. {For each submap~$m$, and its frame set $\mathcal{S}_m$, its anchor-frame set $\mathcal{B}_m\subset\mathcal{S}_m$ contains the first and last $N$ frames, matching the anchor-group size defined above. Because the stage is offline, both groups are available before optimization, and thus constrain the submap at its two ends.}

\paragraph{Seeded Boostrap with Submap Anchors.} {We collect the first $N_{\mathrm{boot}}=8$ keyframes}
of $\mathcal{S}_m$, initialize the {front anchor frames} with their poses obtained from Sec. \ref{sec:anchor},
and keep them \underline{fixed} while refining the remaining bootstrap cameras, sparse
points, and optionally the focal length.  This directly initializes the submap
in the aerial metric frame and avoids a separate Sim(3) alignment step.

\paragraph{Incremental Registration.} Non-anchor frames are then registered incrementally from 2D--3D
correspondences with selected keyframes.  Rather than choosing references only by temporal proximity, we use a 3D-aware keyframe selection that ranks candidates by the number of matched reference keypoints with existing triangulated 3D points.  This approach provide usable PnP constraints and avoids visually similar but geometrically weak references in low-overlap segments.  For each incoming frame, we match dense features~\cite{edstedt2024roma}
to the selected reference keyframes, keep matches whose reference keypoints
have valid 3D points, estimate a pose with PnP-RANSAC, and refine it with a
one-frame {mini-BA} following \cite{meuleman2025onthefly}.  In low-parallax open scenes, we optionally
filter PnP inputs by reference-camera depth and enforce an explicit pixel
reprojection threshold to reduce far-point bias.

\paragraph{Fixed-Anchor Insertion.} Anchor frames are still matched to the local map so they contribute tracks, but
their inserted poses are overwritten by their localized anchor poses. This
is applied to both front and rear anchor groups, making the rear group a true
boundary condition rather than a fallback when PnP fails.

\paragraph{Anchor-Constrained Bundle Adjustment.} After all frames in $\mathcal{S}_m$ are processed, we run a final
{anchor-constrained bundle adjustment (BA)} illustrated in Fig. \ref{fig:anchor_ba} (b) and (c).  All anchor frames are restored to their {localized poses}, and only non-anchor camera poses and local 3D points are optimized.
{Let $\mathcal{O}_m$ be the valid frame--point observation pairs in submap $m$. Each $(i,j)\in\mathcal{O}_m$ associates frame $I_i^g$ with local 3D point $\mathbf{X}_j$ and its measured image position $\mathbf{u}_{ij}$. We use the shared intrinsic matrix $\mathbf{K}$, robust loss $\rho$, and localized anchor pose $\mathbf{T}_{i,\mathrm{anc}}^g$; $\pi(\mathbf{K},\mathbf{T},\mathbf{X})$ projects point $\mathbf{X}$ under pose $\mathbf{T}$ and intrinsics $\mathbf{K}$. We retain anchor-frame observations in the objective through the effective pose $\tilde{\mathbf{T}}_i^g$:}
\begin{equation}
{
\begin{aligned}
\tilde{\mathbf{T}}_i^g
&=
\begin{cases}
\mathbf{T}_i^g, & I_i^g \in \mathcal{S}_m \setminus \mathcal{B}_m,\\
\mathbf{T}_{i,\mathrm{anc}}^g, & I_i^g \in \mathcal{B}_m,
\end{cases}\\
\min_{\substack{\{\mathbf{T}_i^g\}_{I_i^g\in\mathcal{S}_m\setminus\mathcal{B}_m}\\
                  \{\mathbf{X}_j\}}}
&\sum_{(i,j)\in\mathcal{O}_m}
\rho\!\left(\left\|\pi(\mathbf{K},\tilde{\mathbf{T}}_i^g,\mathbf{X}_j)-\mathbf{u}_{ij}\right\|_2^2\right).
\end{aligned}}
\end{equation}
{For a non-anchor frame, $\tilde{\mathbf{T}}_i^g$ is its optimized pose; for an anchor frame, it is the fixed localized pose. Both types of observations therefore contribute to the same reprojection objective without allowing the anchor poses to drift. Tracks shared by anchor and non-anchor frames transfer the metric constraints into the submap interior, and we prioritize those observed by multiple keyframes. The resulting submap trajectories are assembled in the common aerial metric frame and passed to the scene update.}

\begin{figure}[!ht]
    \centering
    \includegraphics[width=1\linewidth]{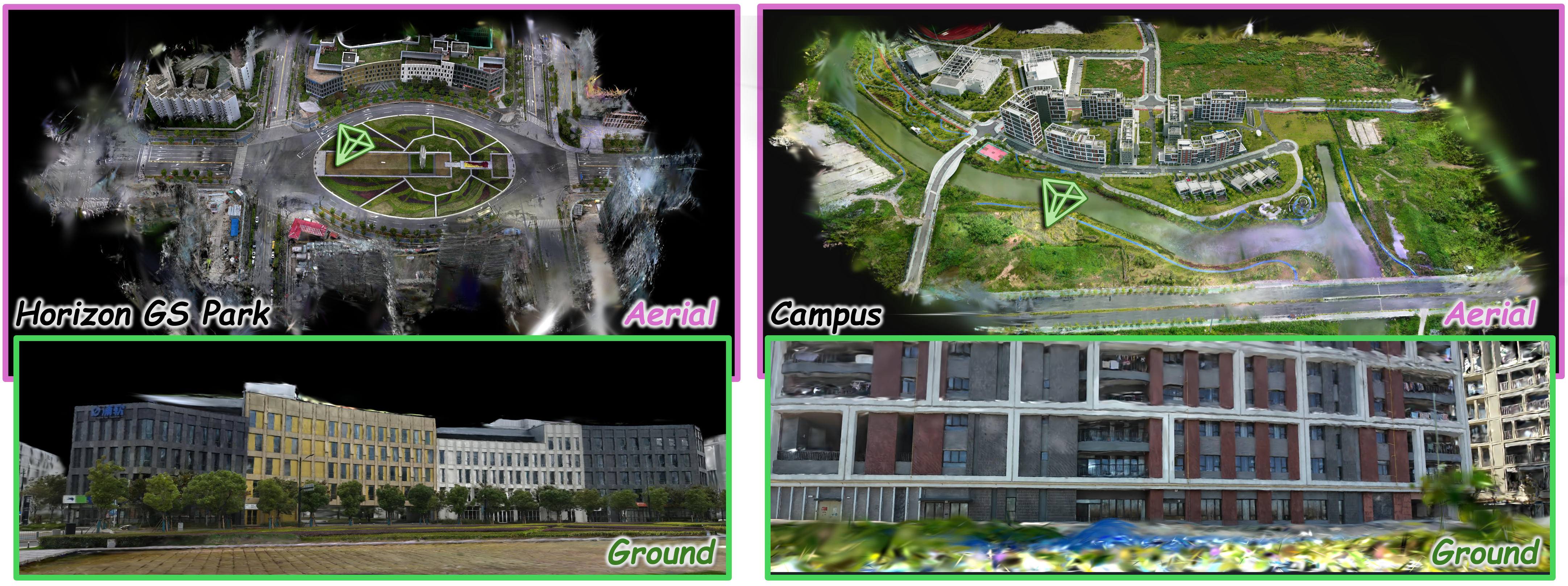}
    \caption{\textbf{{Cross-view scene update examples.}}}
    \label{fig:atdemo}
\end{figure}

\definecolor{groupgray}{RGB}{245,246,248}
\definecolor{rankgreen}{HTML}{59A14F}
\newcommand{\first}[1]{\cellcolor{rankgreen!35}\textbf{#1}}
\newcommand{\second}[1]{\cellcolor{rankgreen!22}#1}
\newcommand{\third}[1]{\cellcolor{rankgreen!12}#1}
\begin{table*}[!ht]
\footnotesize
    \setlength\tabcolsep{1pt}
    \centering 
    \renewcommand\arraystretch{1}
    \caption{\textbf{Quantitative comparison of anchor localization.} We evaluate anchor poses on seven datasets using four metrics: ATE, RPE-t, RPE-r, and registration rate. The top two results for each scene are highlighted in \colorbox{rankgreen!22}{green}. When a reconstruction cannot be aligned to the ground-truth reference, only RPE$_r$ and registration rate are reported.}
    \label{tab:anchor_reconstruction}
    \resizebox{\linewidth}{!}{
\begin{tabular}{l *{28}{c}}
\toprule
\multirow{2}{*}{Method}
& \multicolumn{4}{c}{Ours-Campus}
& \multicolumn{4}{c}{HorizonGS-Park}
& \multicolumn{4}{c}{HorizonGS-Road}
& \multicolumn{4}{c}{ISPRS-Rathaus}
& \multicolumn{4}{c}{ISPRS-Stadthaus}
& \multicolumn{4}{c}{ISPRS-Zeche}
& \multicolumn{4}{c}{ALoG-Uni10k} \\
\cmidrule(lr){2-5}
\cmidrule(lr){6-9}
\cmidrule(lr){10-13}
\cmidrule(lr){14-17}
\cmidrule(lr){18-21}
\cmidrule(lr){22-25}
\cmidrule(lr){26-29}
& ATE$\downarrow$ & RPE-t$\downarrow$ & RPE-r$\downarrow$ & Reg.
& ATE$\downarrow$ & RPE-t$\downarrow$ & RPE-r$\downarrow$ & Reg.
& ATE$\downarrow$ & RPE-t$\downarrow$ & RPE-r$\downarrow$ & Reg.
& ATE$\downarrow$ & RPE-t$\downarrow$ & RPE-r$\downarrow$ & Reg.
& ATE$\downarrow$ & RPE-t$\downarrow$ & RPE-r$\downarrow$ & Reg.
& ATE$\downarrow$ & RPE-t$\downarrow$ & RPE-r$\downarrow$ & Reg. 
& ATE$\downarrow$ & RPE-t$\downarrow$ & RPE-r$\downarrow$ & Reg. \\
\midrule
{COLMAP}      & - & - & - & 0/48 
& - & - & - & 0/108 
& - & - & - & 0/36 
& \first{0.11} & \first{0.02} & \first{0.04} & \first{36/36} 
& \second{0.43} & \first{0.05} & \first{0.06} & 57/60 
& \second{0.22} & \first{0.01} & \first{0.02} & 18/24 
& \second{8.37} & \second{0.43} & \second{0.63} & 12/60 \\
{InstantSfM}     & - & - & - & 0/48 
& \second{90.43} & \second{60.07} & \second{6.55} & 12/108 
& - & - & - & 0/36 
& 4.45 & 3.45 & 7.27 & \first{36/36} 
& 0.70 & 0.34 & 0.15 & \first{60/60} 
& 0.23 & \second{0.01} & \first{0.0} & 18/24 
& - & - & 6.77 & 18/60 \\
{MASt3R-SfM}   & \second{28.45} & \second{3.74} & \second{7.93} & \first{48/48}
& - & - & 9.40 & \first{108/108} 
& \second{71.05} & \second{13.26} & \second{4.68} & \first{36/36} 
& 33.15 & 3.03 & 2.92 & \first{36/36} 
& 15.48 & 0.62 & 0.50 & \first{60/60}
& 12.59 & 2.17 & 4.24 & \first{24/24} 
& 35.61 & 5.32 & 5.11 & \first{60/60} \\
{MP-SfM}       & - & - & - & 0/48 
& - & - & - & 0/108 & - & - & - & 0/36 
& 0.46 & 0.06 & 0.100 & \first{36/36} 
& - & - & - & 0/60 
& - & - & - & 0/24
& 18.53 & 1.56 & 3.12 & \first{60/60} \\
\midrule
Ours                & \first{0.06} & \first{0.02} & \first{0.33} & \first{48/48} 
& \first{0.17} & \first{0.06} & \first{0.16} & \first{108/108} 
& \first{0.10} & \first{0.03} & \first{0.08} & \first{36/36} 
& \second{0.14} & \second{0.02} & \second{0.07} & \first{36/36} 
& \first{0.30} & \second{0.08} & \second{0.10} & \first{60/60} 
& \first{0.17} & 0.018 & \second{0.08} & \first{24/24} 
& \first{0.21} & \first{0.04} & \first{0.21} & \first{60/60} \\
\bottomrule
\end{tabular}}
\end{table*}

\begin{table}[!ht]
\footnotesize
    \setlength\tabcolsep{1pt}
    \centering 
    \renewcommand\arraystretch{1}
    \caption{\textbf{Quantitative comparison of trajectory recovery.} "-" indicates that a method fails or disconnected reconstructions.}
    \label{tab:ground_trajectory_reconstruction}
    \resizebox{\linewidth}{!}{
\begin{tabular}{l *{12}{c}}
\toprule
\multirow{2}{*}{Method}
& \multicolumn{3}{c}{Ours-Campus}
& \multicolumn{3}{c}{HorizonGS-Park}
& \multicolumn{3}{c}{ISPRS-Rathaus}
& \multicolumn{3}{c}{ALoG-Uni10k} \\
\cmidrule(lr){2-4}
\cmidrule(lr){5-7}
\cmidrule(lr){8-10}
\cmidrule(lr){11-13}
& \scriptsize{ATE}$\downarrow$ & \scriptsize{RPE-t}$\downarrow$ & \scriptsize{RPE-r}$\downarrow$
& \scriptsize{ATE}$\downarrow$ & \scriptsize{RPE-t}$\downarrow$ & \scriptsize{RPE-r}$\downarrow$
& \scriptsize{ATE}$\downarrow$ & \scriptsize{RPE-t}$\downarrow$ & \scriptsize{RPE-r}$\downarrow$
& \scriptsize{ATE}$\downarrow$ & \scriptsize{RPE-t}$\downarrow$ & \scriptsize{RPE-r}$\downarrow$ \\
\midrule

\multicolumn{13}{>{\columncolor{groupgray}}l}{\textbf{Structure-from-Motion}} \\
{COLMAP} & - & - & - &  - &  - & - & 0.02&  0.01& 0.04& 8.64 & 3.43 & 5.87      \\
{InstantSfM}  & 10.11 & 1.23 & 10.04 &  38.64 &  13.63 & 21.89 & 0.12&  0.03& 0.01& 26.18& 7.37& 4.58      \\
{MASt3R-SfM}   & 8.49 & 0.34 & 12.21 &  15.13 &  3.65 & 2.72 & 2.54&   0.59 &  0.74& 19.38 & 1.57 & 12.02      \\

\midrule
\multicolumn{13}{>{\columncolor{groupgray}}l}{\textbf{Feed-forward SLAM}} \\

{MASt3R-SLAM}    & 2.44 & 0.82 & 1.21 &  - &  - & - & - & - & - & 10.94 & 3.04& 3.88       \\
{VGGT-SLAM}    & 4.03 & 0.58 & 1.63 & 7.61 &  0.41&  0.69 & 5.53 & 4.07 & 5.17 & 3.83 & 0.15 &0.67   \\

\midrule
\multicolumn{13}{>{\columncolor{groupgray}}l}{\textbf{Simple PnP}} \\
On-the-fly NVS & 4.84 & 0.68 & 0.52 & 6.43 & 0.76 & 0.14 & 24.49 & 5.58 & 6.03& 16.65 & 0.39  & 2.35 \\
Ours (w/o Sim(3))  & 1.40 & 0.61 & 0.46 & 0.19 & 0.03 & 0.03 &  0.19 & 0.026 & 0.06 & 0.23 & 0.65 & 0.40    \\ 

\bottomrule
\end{tabular}}
\end{table}

\subsection{Cross-View Scene Update}
The final stage updates the scene while keeping the aerial Gaussian model
$\mathcal{G} ^a$ as the representation backbone.  With {ground cameras in the aerial metric frame}, we
triangulate and filter a dense ground point cloud, initialize ground Gaussians
$\mathcal{G} ^g$, and append them to the scene without additional alignment.  Let $\mathcal{I}_{\mathrm{tr}}^g\subseteq\mathcal{I}^g$ be the ground training views, let $\mathcal{L}(I;\Gamma)$ be the vanilla 3DGS rendering loss \cite{kerbl2023gaussian}, and let $\mathcal{G}^a\cup\mathcal{G}^g$ denote the concatenated aerial and ground Gaussian sets. During this stage, we optimize only $\mathcal{G} ^g$:
\begin{equation}
{
\min_{\mathcal{G}^g}
\sum_{I\in\mathcal{I}_{\mathrm{tr}}^g}\mathcal{L}\bigl(I;\mathcal{G}^a\cup\mathcal{G}^g\bigr),
\qquad \mathcal{G}^a\ \text{fixed}.}
\end{equation}
{The aerial Gaussians still participate in rendering but receive no gradients, so the optimizable ground Gaussians absorb the ground-view residuals while the aerial model remains unchanged. The fitted $\mathcal{G} ^g$ then initializes joint refinement.}

{During joint refinement, both $\mathcal{G}^a$ and$\mathcal{G} ^g$ are unfrozen and optimized over the combined aerial and ground training views. We use a short schedule with further densification disabled to reduce residual cross-domain inconsistencies. Demonstrated in Fig. \ref{fig:atdemo}, the refined $\mathcal{G}^a\cup\mathcal{G}^g$ is the updated ground-view examples.}

\subsection{{Implementation Details.}}
{We use the same fixed implementation settings across all seven scenes, without scene-specific manual tuning. The aerial visibility graph retains $N_{\mathrm{vis}}=16$ neighbors per image. For each ground frame, we retrieve $N_{\mathrm{ret}}=20$ aerial candidates and return $N_{\mathrm{near}}=3$ graph neighbors for the selected seed. We accept a local reconstruction only when its reprojection error does not exceed $4.0$ pixels. For feature matching, we extract up to 6,000 keypoints per image and run RANSAC with confidence $0.99999$, discarding correspondences with an error above $0.001$ and retaining candidate keyframes only when they produce more than 100 inliers. During anchor-constrained BA, we retain tracks with length greater than three and reprojection error below $2.0$ pixels.}

\section{Experiments}
\subsection{{Experimental Setup.}}

\paragraph{Datasets.}
We evaluate on seven cross-view scenes that contain both aerial observations and ground-view sequences: HorizonGS-Park, HorizonGS-Road \cite{jiang2025horizongs}, ISPRS-Rathaus, ISPRS-Stadthaus, and ISPRS-Zeche \cite{nex2015isprsbenchmark}, ALoG-Uni10k \cite{windisch2026alod}, and our own Ours-Campus capture. For the {HorizonGS scenes}, we \underline{remove} the aerial--ground linking images. These scenes cover real-world environments with different spatial scales, image densities, and reconstruction conventions. Detailed dataset preprocessing is described in Appendix.

\definecolor{rankblue}{HTML}{4C78A8}
\definecolor{rankgreen}{HTML}{59A14F}
\newcommand{\Afirst}[1]{\cellcolor{rankblue!35}\textbf{#1}}
\newcommand{\Asecond}[1]{\cellcolor{rankblue!22}#1}
\newcommand{\Athird}[1]{\cellcolor{rankblue!12}#1}
\newcommand{\Gfirst}[1]{\cellcolor{rankgreen!35}\textbf{#1}}
\newcommand{\Gsecond}[1]{\cellcolor{rankgreen!22}#1}
\newcommand{\Gthird}[1]{\cellcolor{rankgreen!12}#1}
\newcommand{\Abox}[1]{%
  {\setlength{\fboxsep}{1pt}\colorbox{rankblue!25}{#1}}%
}
\newcommand{\Gbox}[1]{%
  {\setlength{\fboxsep}{1pt}\colorbox{rankgreen!25}{#1}}%
}
\begin{table}[!t]
\centering
\footnotesize
\setlength{\tabcolsep}{1pt}
\renewcommand{\arraystretch}{1}
\caption{Quantitative comparison of novel-view synthesis quality across aerial and ground views. The top three \Abox{aerial} and \Gbox{ground} results for each scene and metric are highlighted in \Abox{blue} and \Gbox{green}, respectively. Baselines use reference poses for both aerial and ground views, whereas SkyAnchor uses anchor-constrained trajectory recovery.}
\label{tab:nvs_quality}

\resizebox{\linewidth}{!}{%
\begin{tabular}{@{}llcccccccccccc@{}}
\toprule
\multirow{3}{*}{Method} & \multirow{3}{*}{View}
& \multicolumn{3}{c}{Small Scale}
& \multicolumn{3}{c}{Mid Scale}
& \multicolumn{6}{c}{Large Scale} \\
\cmidrule(lr){3-5}\cmidrule(lr){6-8}\cmidrule(l){9-14}
& & \multicolumn{3}{c}{{ISPRS-Zeche}}
& \multicolumn{3}{c}{{ISPRS-Rathaus}}
& \multicolumn{3}{c}{HorizonGS-Park}
& \multicolumn{3}{c}{Ours-Campus} \\
\cmidrule(lr){3-5}\cmidrule(lr){6-8}\cmidrule(lr){9-11}\cmidrule(l){12-14}
& & \scriptsize{SSIM}$\uparrow$ & \scriptsize{PSNR}$\uparrow$ & \scriptsize{LPIPS}$\downarrow$
& \scriptsize{SSIM}$\uparrow$ & \scriptsize{PSNR}$\uparrow$ & \scriptsize{LPIPS}$\downarrow$
& \scriptsize{SSIM}$\uparrow$ & \scriptsize{PSNR}$\uparrow$ & \scriptsize{LPIPS}$\downarrow$
& \scriptsize{SSIM}$\uparrow$ & \scriptsize{PSNR}$\uparrow$ & \scriptsize{LPIPS}$\downarrow$ \\
\midrule
\multirow{2}{*}{\textbf{PGSR}}
& A (GT)& 0.76 & 20.93 & 0.22 & 0.89 & 27.02 & 0.15 & \Athird{0.84} & 25.09 & \Athird{0.24} & 0.76 & 23.60 & 0.25 \\
& G (GT)& \Gthird{0.87} & \Gthird{25.94} & \Gthird{0.13} & \Gsecond{0.80} & \Gthird{23.73} & \Gsecond{0.20} & \Gthird{0.81} & \Gsecond{25.99} & \Gsecond{0.25} & \Gthird{0.73} & 22.26 & 0.29 \\
\midrule
\multirow{2}{*}{\textbf{GOF}}
& A (GT)& 0.72 & 20.40 & 0.31 & 0.86 & 26.48 & 0.21 & 0.80 & 24.65 & 0.33 & 0.71 & 22.87 & 0.32 \\
& G (GT)& 0.82 & 24.19 & 0.20 & 0.77 & 23.61 & 0.24 & 0.80 & 25.93 & 0.28 & 0.71 & 21.84 & 0.31 \\
\midrule
\multirow{2}{*}{\textbf{Octree-GS}}
& A (GT)& 0.76 & 21.47 & 0.26 & 0.83 & 24.91 & 0.23 & 0.71 & 23.54 & 0.42 & 0.75 & \Athird{24.27} & 0.27 \\
& G (GT)& 0.85 & 24.95 & 0.17 & 0.70 & 21.68 & 0.33 & 0.74 & 23.56 & 0.36 & 0.72 & \Gthird{22.50} & \Gthird{0.29} \\
\midrule
\multirow{2}{*}{\textbf{FLoD}}
& A (GT)& \Athird{0.82} & 23.40 & 0.21 & 0.86 & \Athird{27.17} & 0.21 & 0.80 & \Athird{25.54} & 0.30 & 0.70 & 23.45 & 0.34 \\
& G (GT)& 0.83 & 23.92 & 0.19 & 0.76 & 22.43 & 0.28 & \Gfirst{0.83} & 25.77 & 0.29 & 0.70 & 21.58 & 0.35 \\
\midrule
\multirow{2}{*}{\textbf{EDGS}}
& A (GT)& \Afirst{0.92} & \Asecond{24.97} & \Afirst{0.08} & \Athird{0.90} & 27.11 & \Athird{0.14} & 0.83 & 24.73 & 0.26 & \Athird{0.77} & 23.74 & \Athird{0.24} \\
& G (GT)& 0.84 & 25.82 & \Gsecond{0.11} & 0.72 & 22.90 & \Gthird{0.21} & 0.68 & 21.25 & 0.31 & 0.67 & 21.23 & 0.35 \\
\midrule
\multirow{2}{*}{\textbf{Horizon-GS}}
& A (GT)& 0.80 & \Athird{23.50} & \Athird{0.21} & \Asecond{0.90} & \Afirst{28.86} & \Asecond{0.13} & \Asecond{0.85} & \Asecond{26.28} & \Asecond{0.23} & \Asecond{0.79} & \Asecond{24.98} & \Asecond{0.22} \\
& G (GT) & \Gsecond{0.89} & \Gsecond{26.14} & 0.16 & \Gthird{0.79} & \Gfirst{24.72} & 0.21 & 0.80 & \Gthird{25.95} & \Gthird{0.26} & \Gfirst{0.73} & \Gfirst{23.02} & \Gfirst{0.24} \\
\midrule
\multirow{2}{*}{\textbf{Ours}}
& A (GT) & \Asecond{0.90} & \Afirst{27.21} & \Asecond{0.09} & \Afirst{0.91} & \Asecond{28.37} & \Afirst{0.11} & \Afirst{0.88} & \Afirst{26.33} & \Afirst{0.18} & \Afirst{0.80} & \Afirst{25.00} & \Afirst{0.20} \\
& G (Ours) & \Gfirst{0.90} & \Gfirst{27.95} & \Gfirst{0.08} & \Gfirst{0.83} & \Gsecond{24.68} & \Gfirst{0.14} & \Gsecond{0.82} & \Gfirst{26.20} & \Gfirst{0.21} & \Gsecond{0.73} & \Gsecond{22.62} & \Gsecond{0.25} \\
\bottomrule
\end{tabular}%
}
\end{table}
\paragraph{Setup.}
For anchor localization, we use the original image resolution for every dataset to preserve fine cross-view matching evidence.  We partition each ground-view sequence into local submaps and use the first 6 and last 6  images as the {front and rear anchor groups}. Dataset-specific submap lengths are chosen according to scene scale and image density and are reported in Appendix. We then evaluate trajectory recovery on scenes with {reference ground trajectories}, where methods must recover all ground cameras. For trajectory recovery and {scene update}, all images are resized to a common 1.6K resolution. In the {scene update experiments}, aerial test views are held out at an interval of eight images and evaluated separately from the ground-view test set. All experiments are conducted on an NVIDIA RTX PRO 6000 Blackwell GPU.

\begin{figure*}[!ht]
    \centering
    \includegraphics[width=.95\linewidth]{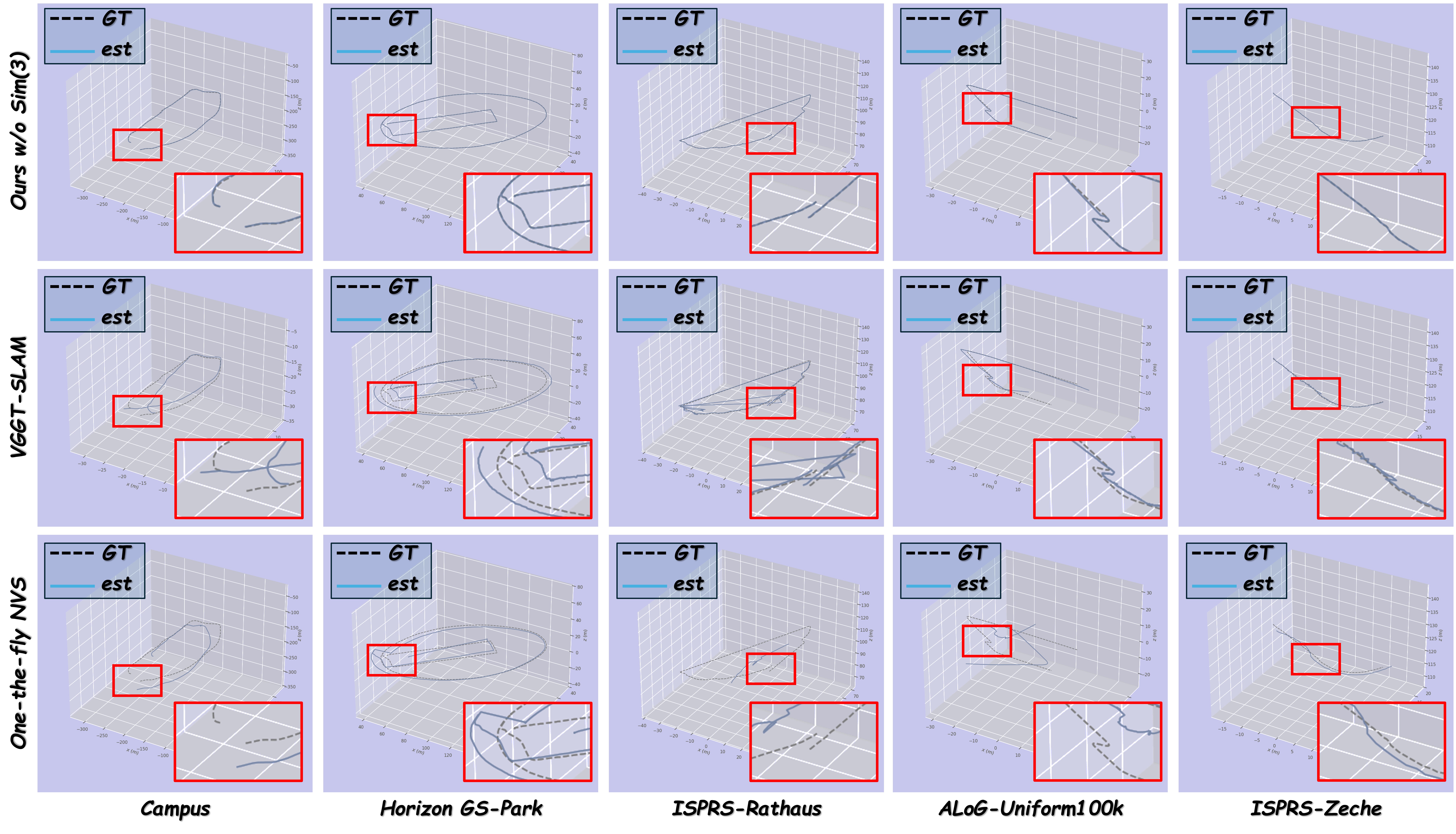}
    \caption{\textbf{Ground trajectory visualization.} We compare the recovered ground trajectories with the reference trajectories for qualitative evaluation. Note that the trajectory from SkyAnchor is evaluated without Sim(3) alignment.}
    \label{fig:Trajcomp}
\end{figure*}
\paragraph{Baselines.}
For anchor localization and trajectory recovery, we compare with COLMAP \cite{schonberger2016sfm}, InstantSfM \cite{zhong2025instantsfm}, MASt3R-SfM \cite{duisterhof2024mast3rsfm}, and MP-SfM \cite{pataki2025mpsfm} as representative large-scale sparse reconstruction pipelines, and include VGGT \cite{wang2025vggt}, VGGT-SLAM \cite{maggio2025vggtslam}, and On-the-fly NVS \cite{meuleman2025onthefly} for trajectory recovery. Methods that do not support reconstruction from a fixed aerial scaffold are run on the joint aerial--ground image set and aligned to the aerial metric frame by a $\mathrm{Sim}(3)$ transformation estimated from aerial camera correspondences \cite{umeyama1991least}; COLMAP is evaluated with its incremental model-extension pipeline \cite{schonberger2016sfm}. Since MASt3R-SfM and MP-SfM both rely on MASt3R-based dense matching \cite{leroy2024mast3r}, we replace their default matcher with Aerial-MASt3R to control the matching backbone. For scene update, we compare with representative Gaussian reconstruction methods, including PGSR \cite{chen2024pgsr}, GOF \cite{yu2024gof}, Octree-GS \cite{ren2024octreegs}, FLoD \cite{seo2024flod}, EDGS \cite{kotovenko2025edgs}, and Horizon-GS \cite{jiang2025horizongs}, adapted to the same aerial--ground input protocol.

\paragraph{Metrics.}
For anchor localization and trajectory recovery, we report ATE, RPE-t, and RPE-r in the aerial metric frame; for anchor localization, we also report the registration rate over anchor frames. When a baseline outputs an arbitrary coordinate system, we align it to the aerial scaffold using reconstructed aerial cameras before evaluation, while \textbf{SkyAnchor is evaluated directly in the aerial metric frame without post-hoc Sim(3) alignment}. We report two {anchor localization failures} separately: {zero registered anchor frames}, where no anchor frame is localized, and \textit{cannot align}, where the $\mathrm{Sim}(3)$ alignment to the aerial scaffold fails. For scene update, we report SSIM, PSNR, and LPIPS on held-out aerial and ground views separately, so that scaffold preservation and street-level reconstruction quality are both visible.

\subsection{Anchor localization}

As shown in Table~\ref{tab:anchor_reconstruction} and Fig.~\ref{fig:anchor_reconstruction_vis}, SkyAnchor registers all anchor frames on all seven scenes and achieves the lowest ATE on six of them. The advantage is most visible on Ours-Campus, HorizonGS-Park, HorizonGS-Road, and ALoG-Uni10k, where several joint-reconstruction baselines either register no anchor frames or produce large alignment errors. COLMAP remains competitive on several ISPRS scenes with stronger geometric overlap, but it fails on the harder large-scale outdoor captures, where high-altitude aerial views introduce severe scale and perspective gaps that make aerial--ground matching much more difficult.

The key observation is that anchor localization cannot be reduced to running a stronger global SfM system on the mixed image set. Baselines such as MASt3R-SfM can register many images, yet their aligned models may still have large metric errors. In contrast, SkyAnchor localizes short anchor groups with verified aerial support, producing sparse but accurate cross-view registration.




\begin{figure*}[!t]
    \centering
    \includegraphics[width=1\linewidth]{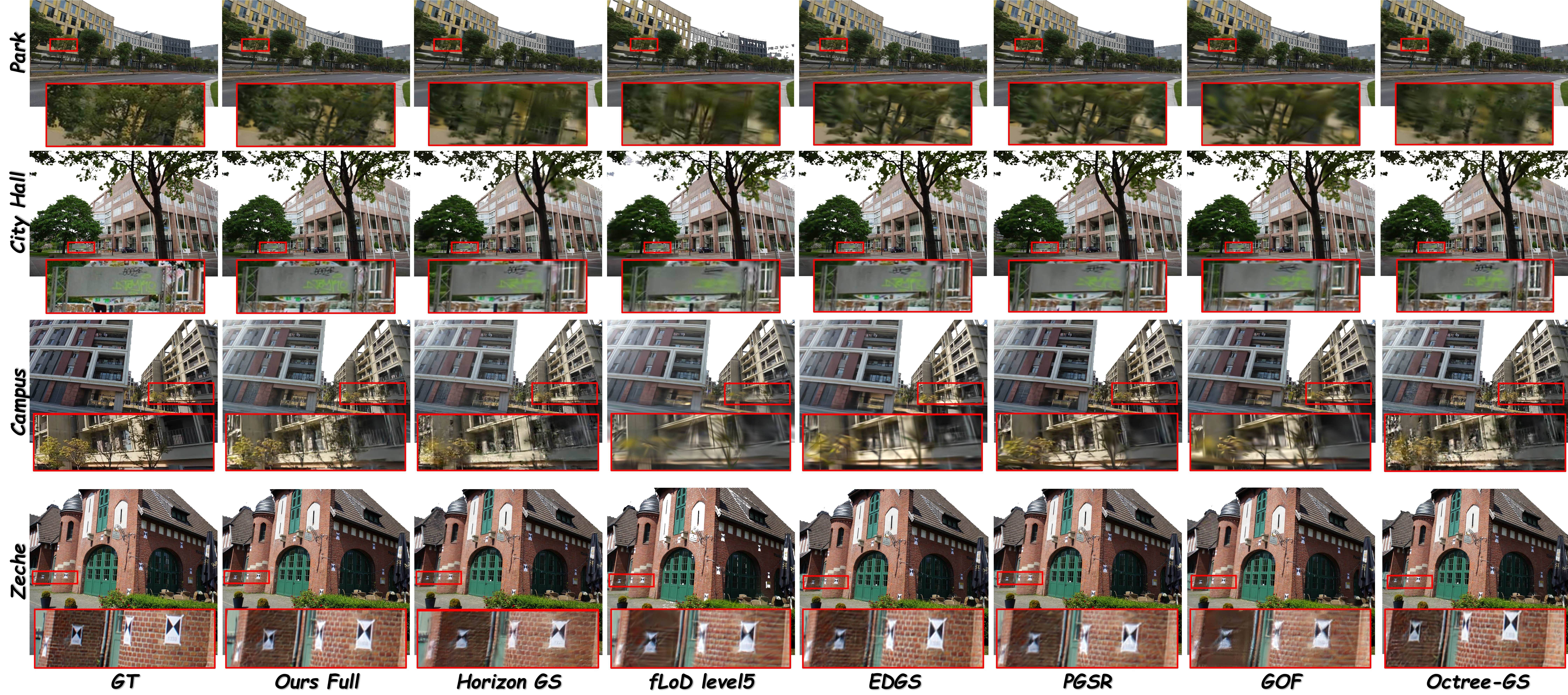}
    \caption{\textbf{Quantitative comparison of {rendering results at ground level}.}}
    \label{fig:groundcomp1}
\end{figure*}
\subsection{Anchor-Constrained Trajectory Recovery}

Table~\ref{tab:ground_trajectory_reconstruction} evaluates whether these sparse anchor poses can stabilize the full ground trajectory. In Fig.~\ref{fig:trajectory_comparison}, SkyAnchor {recovers the trajectory in the aerial metric frame} and improves over On-the-fly NVS \cite{meuleman2025onthefly}, which propagates poses from local 2D--3D registration without persistent anchor constraints. Table~\ref{tab:ground_trajectory_reconstruction} and Fig.~\ref{fig:Trajcomp} show the largest gaps on ISPRS-Rathaus and ALoG-Uni10k, where drift accumulates quickly for the PnP-only baseline but is reduced by anchor-constrained submaps.
The comparison also clarifies the difference from feed-forward reconstruction. Feed-forward methods such as VGGT-SLAM can produce a plausible trajectory, but it is not designed to prevent the trajactory dirfts throughout sequence growth.

\subsection{{Ground-View Update}}

Table~\ref{tab:nvs_quality} evaluates the final update target: preserving aerial rendering quality while adding ground-view detail. The comparisons in Figs.~\ref{fig:groundcomp1} and Table~\ref{tab:nvs_quality} are conservative because the baselines use reference poses for both aerial and ground views, whereas SkyAnchor renders ground views with its recovered trajectory. Even under this setting, SkyAnchor remains competitive on aerial metrics and achieves the best or second-best ground-view quality on most scenes.
 More importantly, the aerial and ground metrics should be read together: improving ground views alone is insufficient if the update damages the {fixed aerial scaffold}. SkyAnchor allows street-level content to be inserted while preserving {aerial scaffold consistency}.

\begin{figure*}[!t]
    \centering
    \includegraphics[width=1\linewidth]{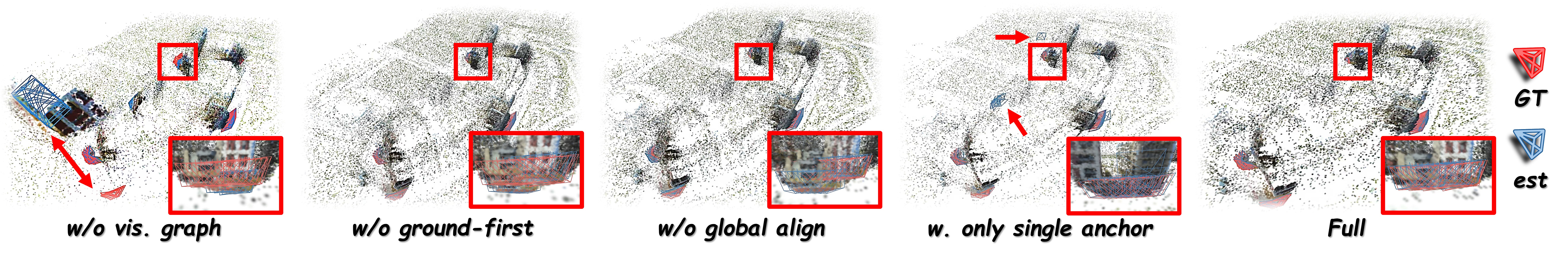}

    \caption{\textbf{{Anchor localization ablations.}} Removing the {aerial visibility graph}, {ground-first reconstruction schedule}, {global metric alignment}, or group-based anchoring degrades anchor localization quality.}
    \label{fig:anchor_ablation}
\end{figure*}
\begin{figure*}[!t]
    \centering

    \includegraphics[width=1\linewidth]{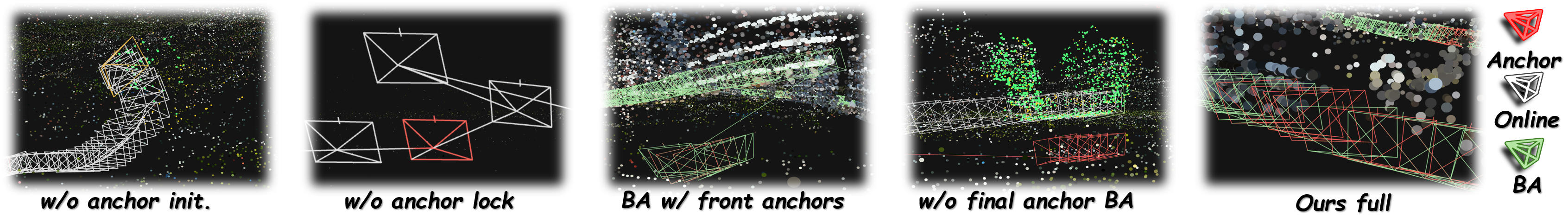}

    \caption{\textbf{Trajectory recovery ablations.} Each component contributes to stable metric propagation.}
    \label{fig:trajectory_ablation}
\end{figure*}

\subsection{Ablation Studies}
\begin{table}[!h]
\centering
\scriptsize
\setlength{\tabcolsep}{4pt}
\renewcommand{\arraystretch}{1}
\caption{\textbf{Anchor localization ablation study.}}

\label{tab:ablation_anchor_avg}
\resizebox{\columnwidth}{!}{
\begin{tabular}{@{}llcccc@{\hspace{9pt}}}
\toprule
 &Variant & ATE$\downarrow$ & RPE-t$\downarrow$ & RPE-r$\downarrow$ & Reg. \\
\midrule
\multirow{5}{*}{Anchor} 
&{w/o aerial vis. graph} & 1.079 & 0.318 & 0.987 & 102/108 \\
&{w/o ground-first recon.} & 0.187 & 0.041 & 0.183 & \first{108/108} \\
&{w/o global metric align.} & 0.310 & 0.070 & 0.300 & \first{108/108} \\
&{single-frame localization} & 1.353 & 1.412 & 8.347 & 97/108 \\
\cmidrule(lr){2-6}
&ours full & \first{0.110} & \first{0.024} & \first{0.163} & \first{108/108} \\
\bottomrule
\end{tabular}}

\end{table}
\paragraph{Anchor Localization Ablations.}
Table~\ref{tab:ablation_anchor_avg} reports anchor localization ablations averaged over {ISPRS-Zeche}, HorizonGS-Road, and Ours-Campus and Fig. \ref{fig:anchor_ablation} provides the visualizations. The full model obtains the best average pose accuracy and complete registration. Removing the {aerial visibility graph} hurts both accuracy and coverage, indicating that retrieval alone does not provide a sufficiently reliable aerial support set. The single-frame variant has the largest rotational error and lower registration, which supports the use of short anchor groups rather than isolated frames. The {variants without the ground-first reconstruction schedule or global metric alignment} still register all anchors but have higher ATE and RPE, showing that registration rate alone is not enough; the anchor poses must also be metrically accurate for downstream trajectory recovery.

\begin{table}[!h]
\centering
\scriptsize
\setlength{\tabcolsep}{4pt}
\renewcommand{\arraystretch}{1}
\caption{\textbf{Trajectory recovery ablation.}}

\label{tab:ablation_traj_avg}
\resizebox{\columnwidth}{!}{
\begin{tabular}{@{}llccc@{\hspace{14pt}}}
\toprule
 & Variant & ATE$\downarrow$ & RPE-t$\downarrow$ & RPE-r$\downarrow$ \\
\midrule
\multirow{7}{*}{Trajectory } 
&w/o anchor init & 22.030 & 3.279 & 4.244 \\
&w/o fixed anchor poses & 4.517 & 1.109 & 1.469 \\
&{front-anchor-only BA} & 1.229 & 1.066 & 0.896 \\
&{w/o anchor-constrained BA} & 3.444 & 2.184 & 2.449 \\
&w/o 3D-aware KF selection & 1.261 & 0.670 & 0.475 \\
\cmidrule(lr){2-5}
&ours full & \first{1.176} & \first{0.510} & \first{0.306} \\
\bottomrule
\end{tabular}}
\end{table}

\paragraph{Trajectory Recovery Ablations.}
Table~\ref{tab:ablation_traj_avg} reports trajectory recovery ablations averaged over the same three scenes. Figure~\ref{fig:trajectory_ablation} shows that the largest failure occurs without anchor initialization, confirming that a ground-only pose chain is not sufficient to enter the aerial metric frame. Allowing anchor poses to move or removing final anchor-constrained BA also increases errors substantially, showing that anchor poses must remain fixed after initialization. The front-anchor-only BA variant keeps ATE close to the full model but worsens local consistency, indicating that {rear anchor poses} help constrain the interior of each submap. Finally, removing 3D-aware keyframe selection gives the smallest degradation, but all metrics still worsen, suggesting that better local support selection improves but does not replace anchor constraints.
\paragraph{Scene Update Ablations.}
Table~\ref{tab:ablation_update_avg} studies which parts of the {scene update} stage matter for ground-view rendering. The aerial-only model performs poorly because it lacks street-level observations, while training from scratch improves ground views but discards the benefit of starting from the {fixed aerial scaffold}. Removing the frozen-aerial insertion stage causes a large drop, showing that ground Gaussians should first explain ground evidence without changing the aerial Gaussian model. {Dense-point initialization from anchor-constrained BA} and final joint refinement further improve the result, with the full staged update achieving the best results.

\paragraph{{Full Cross-View Comparison Against SfM Solutions.}}
{Table~\ref{tab:additional_results} and Fig.~\ref{fig:staff_full_comp} compares full aerial--ground reconstruction on Ours-Campus, containing 419 aerial and 436 ground images. Under our protocol, ground-image GPS priors, low-altitude aerial linking images, and manual tie points are unavailable, causing COLMAP and MP-SfM with MASt3R to fail at aerial--ground registration. Using the same Aerial-MASt3R matcher enables MP-SfM to reconstruct the scene, but SkyAnchor reduces the global ATE from $5.541$ to $1.404$ and improves the 3DGS PSNR from $19.13$ to $22.62$ dB. MP-SfM yields lower local RPE-t/RPE-r, whereas SkyAnchor provides substantially better global alignment and rendering quality.}

\begin{table}[!t]
\centering
\scriptsize
\setlength{\tabcolsep}{4pt}
\renewcommand{\arraystretch}{1}
\caption{\textbf{Ground-view scene update ablation.}}
\label{tab:ablation_update_avg}
\resizebox{\columnwidth}{!}{
\begin{tabular}{@{}llccc@{\hspace{14pt}}}
\toprule
 & Variant & SSIM$\uparrow$ & PSNR$\uparrow$ & LPIPS$\downarrow$ \\
\midrule
\multirow{6}{*}{NVS } 
& aerial only & 0.514 & 16.24 & 0.51   \\
& train from scratch & 0.687 & 22.01 & 0.39   \\
&{w/o frozen-aerial } & 0.529  & 17.16  & 0.50 \\
&{w/o anchor-constrained BA init.} &0.774 & 23.19 & 0.24 \\
&{w/o joint refinement} & 0.784 & 24.46  & 0.21  \\
\cmidrule(lr){2-5}
&ours full & \first{0.816} & \first{25.59} & \first{0.18} \\
\bottomrule
\end{tabular}}
\end{table}

\begin{table}[!t]
\footnotesize
    \setlength\tabcolsep{1pt} 
    \setlength{\tabcolsep}{4pt}
    \renewcommand\arraystretch{1.2}
    \caption{\textbf{Full scene reconstruction.}}

    \label{tab:additional_results}
    \resizebox{\linewidth}{!}{
\begin{tabular}{lccccc}
\toprule
            \multirow{2}{*}{Method} & \multicolumn{5}{c}{Ours-Campus} \\ \cline{2-6} 
             & ATE & RPE-T & RPE-R & PSNR & Time \\ \hline
            \multicolumn{1}{l}{COLMAP} & fail & fail & fail & fail & - \\
            \multicolumn{1}{l}{MP-SfM(MASt3R)} & fail & fail &fail & fail & - \\
            \multicolumn{1}{l}{MP-SfM(aerial-MASt3R)} & 5.541 & 0.143 & 0.236 & 19.13 & 26h26min \\
            \multicolumn{1}{l}{Ours(aerial-MASt3R)} & 1.404 & 0.598 & 0.457 & 22.62 & 3h18min \\
\bottomrule
\end{tabular}}
\end{table}
\begin{figure}[!t]
    \centering
    \includegraphics[width=1\linewidth]{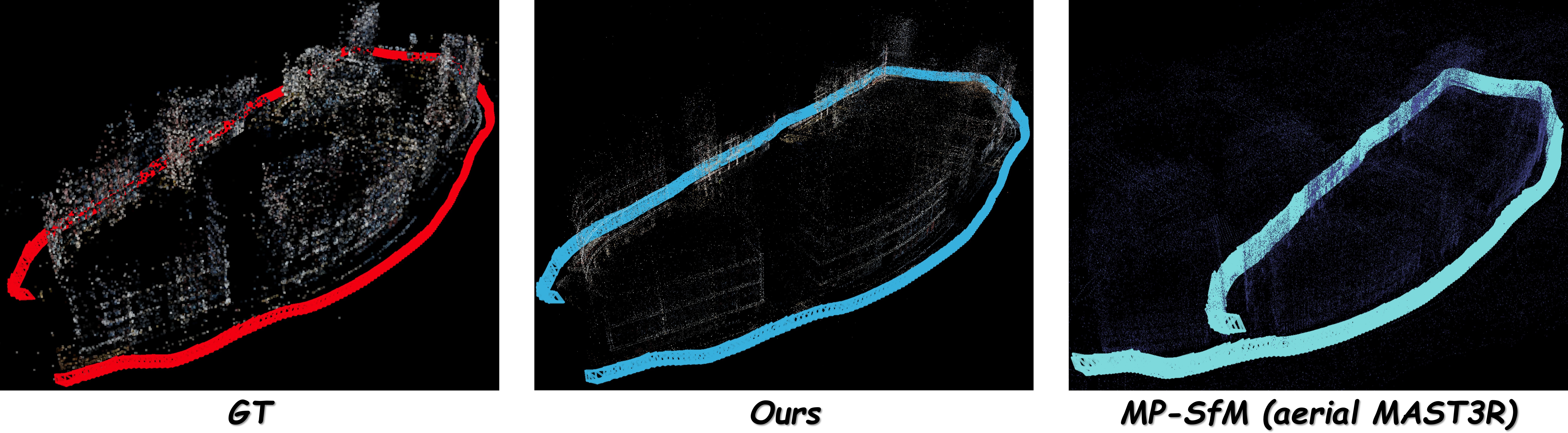}

    \caption{\textbf{Full joint cross-view reconstruction comparison.}}
 
    \label{fig:staff_full_comp}
\end{figure}

\begin{figure*}[!t]
    \centering

    \begin{minipage}[t]{0.44\textwidth}
        \centering
        \includegraphics[width=\linewidth]{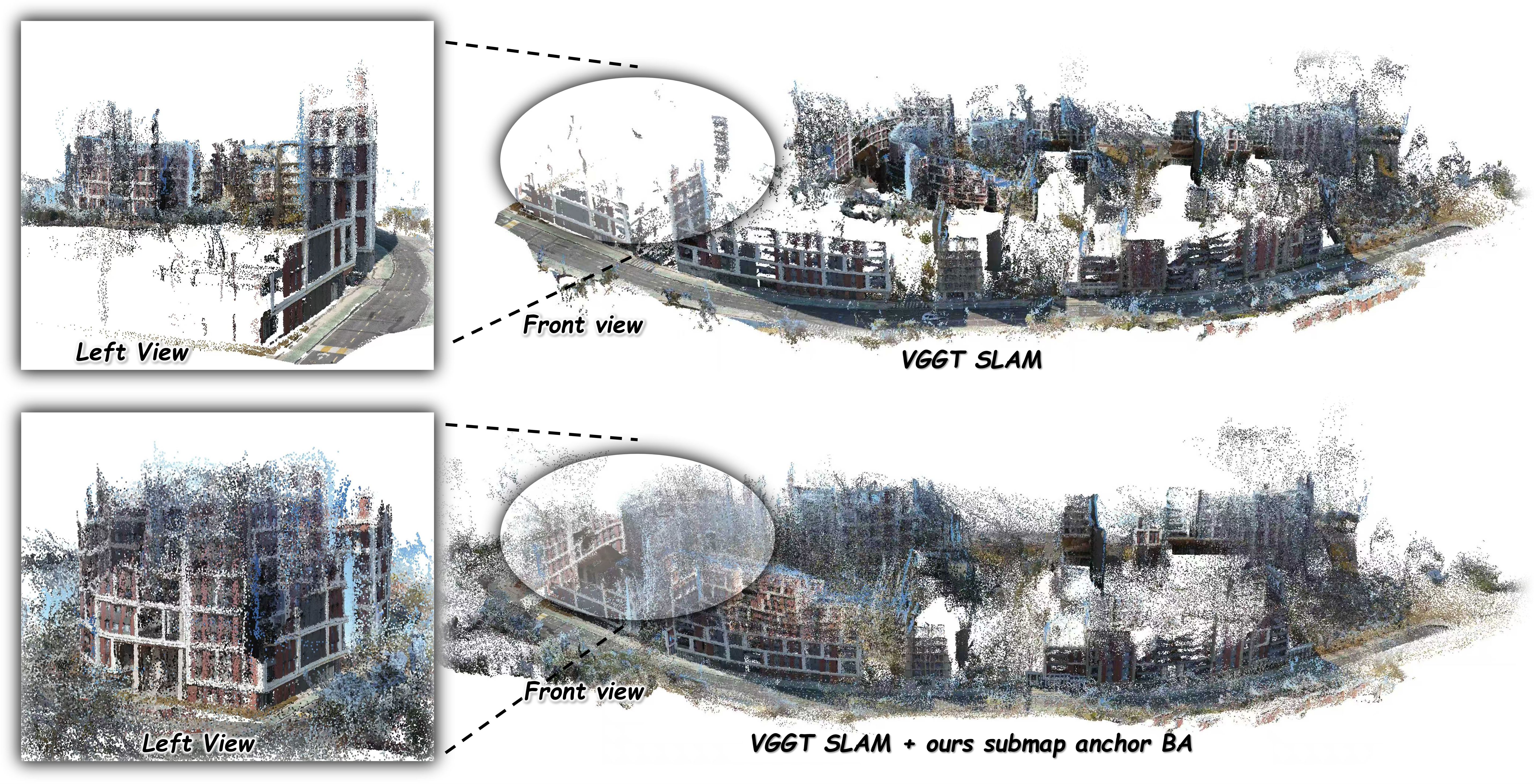}

        \centerline{\small  (a) Effectiveness of anchor-constrained BA.}
    \end{minipage}
    \hfill
    \begin{minipage}[t]{0.48\textwidth}
        \centering
        \includegraphics[width=\linewidth]{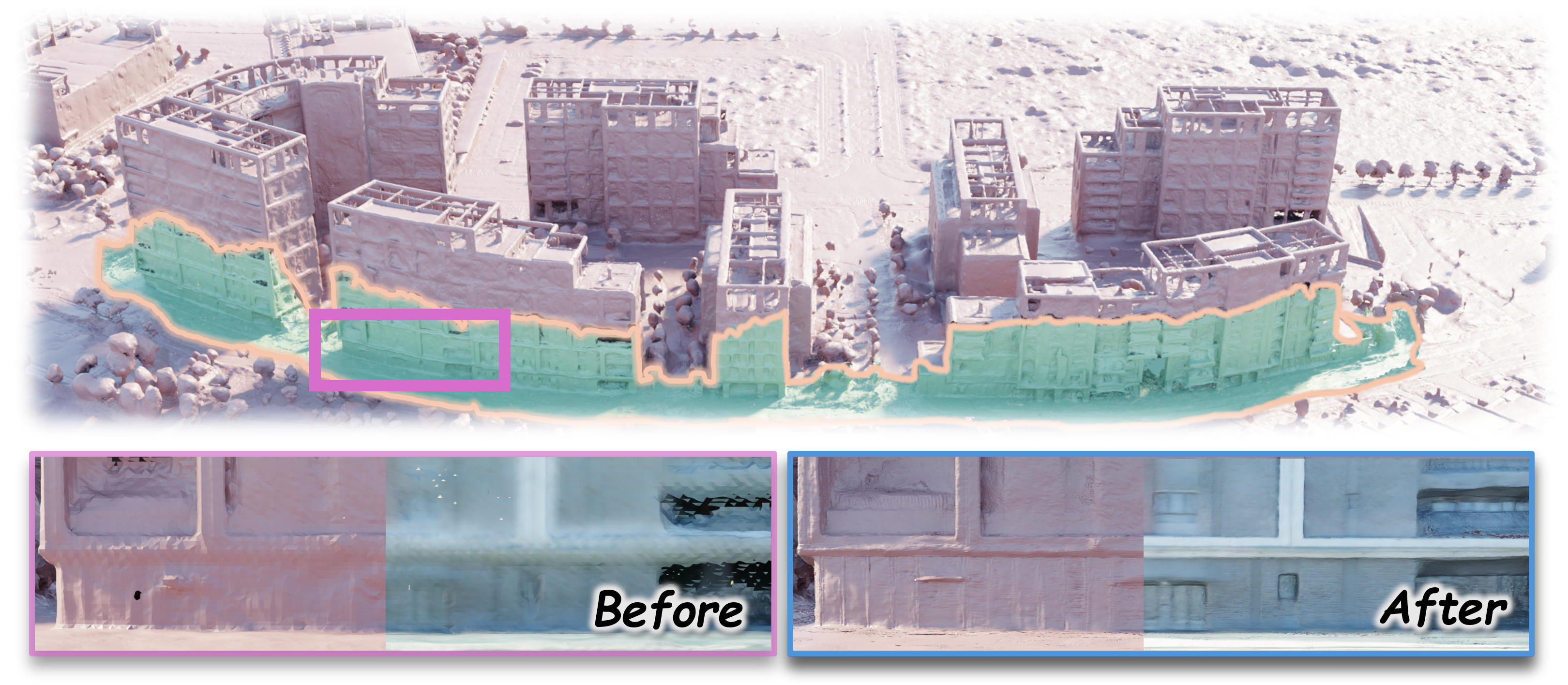}

        \centerline{\small (b) Ground Mesh update.}
    \end{minipage}

    \caption{
    \textbf{Left}: Unconstrained feed-forward methods tend to suffer from severe drift and scale errors as the sequence progresses (e.g., in the latter part of the {Ours-Campus} scene). Our anchor-constrained bundle adjustment effectively mitigates these accumulative errors, ensuring a strictly georeferenced and metrically accurate reconstruction.
    \textbf{Right}: Apply SkyAnchor on 2DGS improves the quality of the ground geometry.
    }
    \label{fig:additional}

\end{figure*}
\begin{figure}[!t]
    \centering
    \includegraphics[width=1\linewidth]{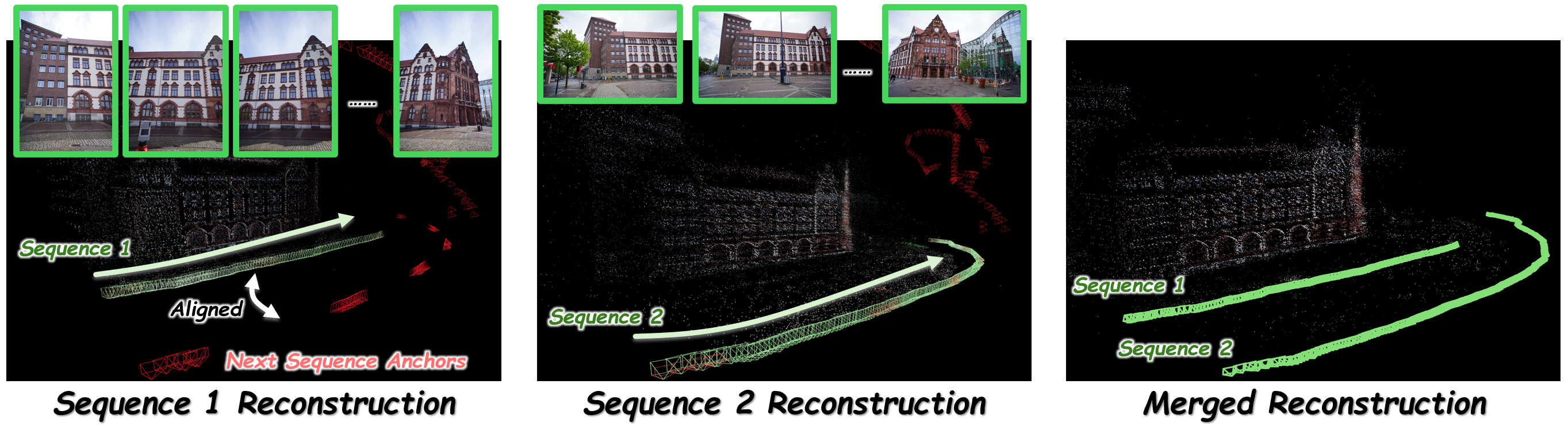}

    \caption{\textbf{{Ground-view maintenance on ISPRS-Stadthaus.}} {Two newly localized anchor groups place an additional ground sequence in the same aerial metric frame as the existing sequence, enabling their trajectories to be merged without joint re-reconstruction.}}

    \label{fig:sequencemerge}
\end{figure}

\paragraph{{Efficiency Analysis.}}
{Table~\ref{tab:additional_results} also shows the efficiency advantage of updating a fixed aerial scaffold over jointly reconstructing all aerial and ground images. MP-SfM with Aerial-MASt3R requires 26~h~26~min, whereas SkyAnchor spends 43~min on cross-view anchor localization and 35~min on ground trajectory recovery. Because SkyAnchor assumes that aerial poses are already available, these 78~min constitute its pose-update cost; the additional 2~h COLMAP aerial reconstruction included in the reported 3~h~18~min total serves only as a proxy for obtaining the assumed aerial poses. Peak GPU memory consumption is 16.9~GB for anchor localization, 13.4~GB for ground trajectory recovery, and 23.6~GB for 3DGS scene update.}

\section{Discussion and Additional Results}
This section presents additional exploratory experiments. Since these results are not rigorously validated, we provide only a qualitative discussion rather than treating them as main evidence.

\paragraph{Applying SkyAnchor to Feed-Forward Methods.}
SkyAnchor can also be extended to feed-forward reconstruction backbones. We first use VGGT-SLAM \cite{maggio2025vggtslam} to reconstruct all ground submaps independently. For each submap, we roughly align its front and back frames to the existing anchors with a Sim(3) transformation, then back-project the dense points onto the neighborhood RGB frames to obtain presudo image correspondences. These correspondences are further optimized with the same anchor-constrained BA used in SkyAnchor. Fig.~\ref{fig:additional}(a) shows that this extension improves the trajectory consistency and point-cloud quality of feed-forward methods to some extent, indicating that our anchoring strategy is not limited to a specific local reconstruction backend.

\paragraph{Ground-View Geometry Extraction.}
To better extract street-level geometry, we replace 3DGS with 2DGS and jointly reconstruct the ground-side scene using the ground-view poses recovered by SkyAnchor. We then estimate a coarse geometry extraction and cropping volume from the ground camera frusta. Specifically, for each ground view, we retain only regions within the camera frustum and discard geometry with depth larger than 20m, which removes distant and weakly constrained structures. As shown by the green regions in Fig.~\ref{fig:additional}(b), this ground-view extraction strategy substantially improves local geometric details, producing sharper and more complete street-level structures.

\paragraph{{Ground-View Maintenance.}}
{SkyAnchor can incorporate a later ground-view sequence without jointly reconstructing it with previously registered ground observations. To evaluate this maintenance setting, we select an additional ground sequence from ISPRS-Stadthaus and localize two new anchor groups against the unchanged aerial scaffold. These anchors place the new sequence directly in the same aerial-view, after which its trajectory is recovered independently and merged with the existing ground reconstruction. As shown in Fig.~\ref{fig:sequencemerge}, the two reconstructed trajectories remain consistent in their overlapping region while preserving the existing reconstruction.}

\section{Conclusion}
We presented SkyAnchor, {a pipeline for updating a fixed aerial scaffold} with a newly captured unposed ground-view sequence. By localizing short {anchor groups}, propagating their metric constraints through anchor-constrained submaps, and updating the aerial Gaussian model while preserving the aerial scaffold, SkyAnchor recovers the ground trajectory in the aerial metric frame and adds street-level appearance and geometry without rerunning mixed-view pose reconstruction.


\paragraph{Failure Cases.} SkyAnchor can still exhibit small residual displacement in regions where aerial and ground views have limited overlap. 

\begin{acks}
This work was supported by the Young Scientists Fund of the National Natural Science Foundation of China (Grant No. 42401567), Guangdong Provincial Project	(Grant No. 2024QN11G095), 
the Open Fund of the State Key Laboratory of Spatial Datum	(Grant No. SKLSD2026-KF-25), and AI Research and Learning Base of Urban Culture  (Grant No. 2023WZJD008).
\end{acks}


\bibliographystyle{ACM-Reference-Format}
\bibliography{reference}

\appendix

\section*{Supplementary Material Overview}
\begin{itemize}
    \item \textbf{Appendix~\ref{app:datasets_protocol}: Datasets and evaluation protocol.} We describe the seven aerial--ground scenes, anchor selection settings, the update-task conversion, and the Gaussian baseline training protocol.
    \item \textbf{Appendix~\ref{app:anchor_localization_details}: Anchor localization details.} We provide the {anchor localization algorithm}, component ablations for retrieval and dense matching, and runtime analysis for anchor localization.
    \item \textbf{Appendix~\ref{app:trajectory_details}: Anchor-constrained trajectory recovery details.} We provide the {trajectory recovery algorithm}, per-submap runtime breakdown, and the {anchor group size} ablation.
\end{itemize}

\section{Datasets and Evaluation Protocol}
\label{app:datasets_protocol}

\subsection{Dataset Details}

{We evaluate seven aerial--ground scenes spanning different spatial scales, capture platforms, and sampling densities (Fig.~\ref{fig:dataset}). We classify ISPRS-Rathaus, ISPRS-Stadthaus, and ISPRS-Zeche as Easy because they retain aerial--ground linking views and strong multi-platform overlap. The remaining four scenes are Hard because they lack linking views or exhibit larger viewpoint, scale, occlusion, or coverage gaps.}
\begin{figure*}[t]
    \centering
    \includegraphics[width=0.9\linewidth]{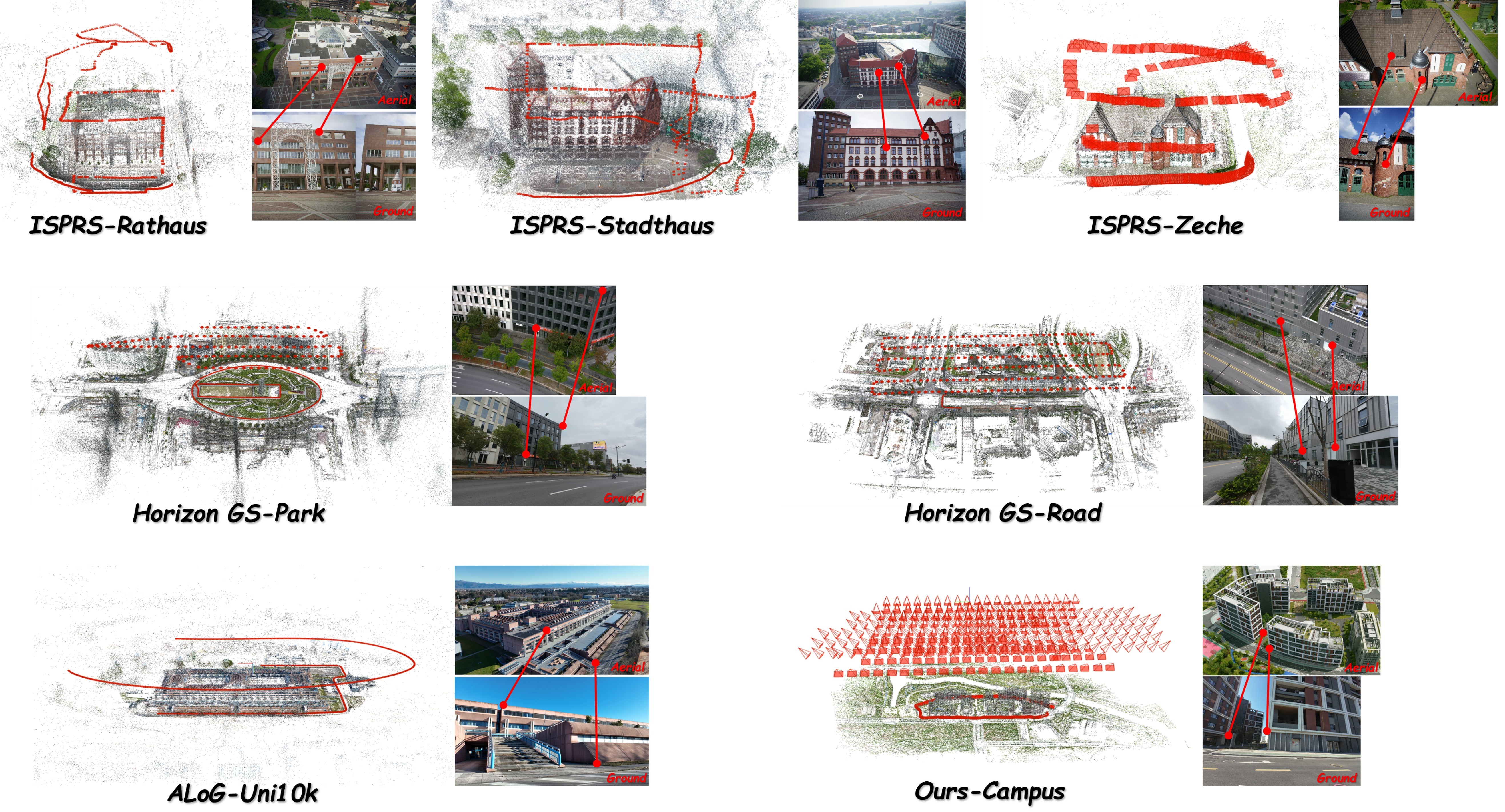}
    \caption{Ground-truth reference results across all datasets.}
    {}
    \label{fig:dataset}
\end{figure*}
\paragraph{ISPRS-Rathaus (Easy).}
{Adapted from the Rathaus scene in the ISPRS multi-platform photogrammetry benchmark, this scene uses 445 aerial images captured by a DJI S800 UAV with a Sony Nex-7 camera in nadir and oblique views, together with 394 ground images. Its structured urban setting contains façades, windows, rooflines, corners, and surrounding square areas; the relatively low flight altitude and limited ground occlusion provide strong cross-view overlap.}

\paragraph{ISPRS-Stadthaus (Easy).}
{ISPRS-Stadthaus follows the same acquisition protocol as ISPRS-Rathaus and contains 345 aerial images and 263 ground images. The scene contains regular planar walls, windows, entrances, and roofs; repeated façade patterns introduce ambiguity despite the strong cross-view overlap.}

\paragraph{ISPRS-Zeche (Easy).}
{Adapted from the Zeche Zollern area, ISPRS-Zeche combines oblique Sony Nex-7 images from a multi-rotor UAV with nadir Panasonic GX-1 images from a fixed-wing platform. We use 116 aerial images and 106 ground images of this historical-building environment, where the linking views provide strong multi-platform overlap.}

\paragraph{HorizonGS-Park (Hard).}
{HorizonGS-Park uses 690 aerial images from a high-altitude, five-camera DJI capture after the transitional images are removed, and 885 ground images from Cam~5 of a six-camera Action4 helmet rig. The scene includes parkland, trees, walkways, buildings, small pedestrian-scale structures, and moving vehicles. Without linking views, tree occlusion and the large aerial--ground viewpoint and appearance gaps make reconstruction difficult.}

\paragraph{HorizonGS-Road (Hard).}
{HorizonGS-Road follows the same acquisition and linking-view removal protocol as HorizonGS-Park. We use 1225 aerial images and 158 ground images from Cam~1. Its elongated road geometry, repetitive roadside structures, and tree occlusions compound the large aerial--ground viewpoint gap, making cross-view matching and trajectory attachment challenging.}

\paragraph{ALoG-Uni10k (Hard).}
{ALoG-Uni10k is a large-scale campus scene captured as single-camera streaming video sequences. We pair the ground-view sequence with the high-altitude aerial sequence having the largest spatial overlap, yielding 512 aerial images and 1022 ground images. Sparse architectural textures, large uniform-color regions, and the aerial--ground viewpoint gap make reconstruction difficult.}

\paragraph{Ours-Campus (Hard).}
{Ours-Campus is collected for this work using a five-camera drone at very high altitude and a single-camera streaming ground-view sequence, providing 419 aerial images and 436 ground images. It covers campus buildings, roads, walkways, vegetation, and open spaces. The highest aerial altitude among our scenes creates the largest scale difference and parallax, while the narrow ground-view coverage, absence of linking images, and ground-level occlusions further increase the difficulty.}

For all seven scenes, the datasets provide sparse reconstruction results in COLMAP format, which we use as the reference ground truth for evaluation.

\subsection{Anchor Selection and Update Protocol}
\label{app:anchor_selection}

\begin{table}[t]
\footnotesize
\setlength\tabcolsep{5pt}
\centering 
\renewcommand\arraystretch{1}
\caption{Anchor selection settings for different datasets.}

\label{tab:anchor_selection}
\begin{tabular}{l *{4}{c}}
\toprule
Dataset & \#Ground & Submap & Group Size & \#Anchors \\
\midrule
Ours-Campus & 436 & 100 & 6 & 48 \\
HorizonGS-Park & 885 & 100 & 6 & 108 \\
HorizonGS-Road & 158 & 50 & 6 & 36 \\
ISPRS-Rathaus & 394 & line-based & 6 & 36 \\
ISPRS-Stadthaus & 263 & 50 & 6 & 60 \\
ISPRS-Zeche & 104 & 50 & 6 & 24 \\
ALoG-Uni10k & 1022 & 210 & 6 & 60 \\
\bottomrule
\end{tabular}
\end{table}

\noindent{We convert each dataset to the update setting by removing its ground camera poses and treating the aerial SfM reconstruction and aerial Gaussian model as the fixed aerial scaffold. If no pre-trained aerial Gaussian model is available, we train it from the reference aerial reconstruction as preprocessing outside the scene update pipeline. Anchor localization uses the original image resolution, whereas trajectory recovery and scene update use 1.6K images; every eighth aerial image is held out for scene update evaluation. For all scenes except ISPRS-Rathaus, we partition the ordered ground-view sequence into submaps and use the first and last six images as the front and rear anchor groups. Because ISPRS-Rathaus contains disconnected ground segments, we instead select three representative segments and apply the same boundary-group rule. Table~\ref{tab:anchor_selection} summarizes these settings.}

\subsection{{Scene Update Baseline Protocol}}
\label{app:baseline_protocol}
{Because the Gaussian baselines assume posed input, we give them reference poses for both aerial and ground training images, whereas SkyAnchor uses the ground trajectory recovered by our method. All methods share the same image resolution, train/test split, and aerial/ground metrics. We preserve each baseline's losses and optimization strategy, changing only data formatting, camera conversion, and chunking required by the common scenes and memory limits.}

\begin{table}[t]
\footnotesize
\setlength\tabcolsep{4pt}
\centering
\renewcommand\arraystretch{1.1}
\caption{Gaussian optimization schedules used for {scene update} comparison. All methods use the same total budget of 100K iterations.}

\label{tab:gaussian_training_protocol}
\begin{tabular}{lccc}
\toprule
Method & \makecell{{Scaffold / Joint}\\{Initialization}} & \makecell{{Frozen-Aerial}\\{Insertion}} & \makecell{{Joint}\\{Refinement}} \\
\midrule
SkyAnchor & 30K aerial-only & 30K ground-only & 40K \\
Baselines & 60K aerial--ground joint & -- & 40K \\
\bottomrule
\end{tabular}
\end{table}

\noindent\textbf{Gaussian optimization protocol.}
{All methods follow the 100K-iteration schedules in Table~\ref{tab:gaussian_training_protocol}. SkyAnchor uses 30K aerial-only iterations to initialize the aerial model when needed, 30K frozen-aerial iterations to fit the inserted ground Gaussians without changing that model, and 40K joint-refinement iterations over the combined training views. Baselines instead use 60K iterations of joint aerial--ground initialization followed by 40K iterations of joint refinement.}

\noindent\textbf{Baseline densification settings.}
{During baseline joint initialization, densification runs every 300 iterations from iteration 2K to 40K, with opacity reset every 6K iterations. It is disabled throughout the 40K joint refinement. All other losses and representation-specific operations remain unchanged, following the Horizon-GS-style large-scene training protocol.}

\section{Anchor Localization Details}
\label{app:anchor_localization_details}
\subsection{{Hyperparameter Settings}}
{Table~\ref{tab:hyperparameters} summarizes the key hyperparameter settings used for cross-view anchor alignment and anchor-constrained trajectory recovery in all experiments.}

\begin{table*}[!t]
\footnotesize
\setlength{\tabcolsep}{6pt}
\centering
\renewcommand{\arraystretch}{1.3}
\caption{{\textbf{Hyperparameter settings.}}}
{}
\label{tab:hyperparameters}
\begin{tabular}{
    p{0.30\linewidth}
    >{\centering\arraybackslash}p{0.10\linewidth}
    p{0.52\linewidth}
}
\toprule
\textbf{Hyperparameter}
& \textbf{Setting}
& \textbf{Description} \\
\midrule
\multicolumn{3}{l}{\textbf{Cross-View Anchor Alignment}} \\

Retrieval top-$K$ ($N_{\mathrm{ret}}$)
& 20
& Number of retrieved candidate images. \\

Visibility neighbors ($N_{\mathrm{near}}$)
& 3
& Number of neighboring images used to construct the visibility graph. \\

RANSAC confidence
& 0.9999
& Confidence level used for RANSAC-based geometric verification. \\

Fragment reprojection error
& $\leq 4.0$ px
& Maximum reprojection error for accepting reconstructed fragments. \\

Minimum two-view inliers
& $\geq 30$
& Minimum number of geometrically verified correspondences required to accept an image pair. \\

RANSAC inlier ratio
& $\geq 0.30$
& Minimum inlier ratio assumed during two-view geometric verification. \\

RANSAC trial range
& $200$--$20{,}000$
& Minimum and maximum numbers of RANSAC trials for two-view geometric verification. \\

Two-view RANSAC reprojection error
& $\leq 2.0$ px
& Maximum reprojection error for classifying cross-view correspondences as geometric inliers. \\

Minimum common aerial views
& $\geq 4$
& Minimum number of shared aerial cameras required for estimating the similarity transformation. \\
\midrule
\multicolumn{3}{l}{\textbf{Anchor-Constrained Trajectory Recovery}} \\

Number of keypoints
& 6000
& Maximum number of  keypoints extracted per image. \\

Match-filter threshold
& 0.01
& Filtering threshold applied to tentative  correspondences. \\

MAGSAC settings
& $1.0$ px / $0.999999$ / $50{,}000$
& Reprojection threshold, confidence, and maximum iterations used for pairwise geometric verification. \\

Online triangulation reprojection error
& $\max(0.001W,1.5\ \mathrm{px})$
& Maximum reprojection error for online triangulation, where $W$ is the processed image width; this equals $1.5$ px in our experiments. \\

PnP-RANSAC settings
& $8000$ / $3.0$ px
& Number of P4P hypotheses and reprojection threshold used for camera pose estimation. \\

Minimum pose inliers
& $\geq 30$
& Minimum number of PnP or incremental MiniBA inliers required to accept a camera pose. \\

Bootstrap MiniBA
& $8$ / $6000$ / $350$
& Numbers of bootstrap keyframes, optimized feature tracks, and solver iterations; each full bootstrap contains six fixed anchors and two free cameras. \\

Incremental reference keyframes
& $4$ ($6$ on retry)
& Number of 3D-ranked reference keyframes selected from a six-frame candidate window. \\

Incremental MiniBA
& $6000$ / $300$
& Maximum number of feature tracks and optimization iterations used for each incremental pose update. \\

Online anchor BA
& Every $10$ frames; $6000$ / $100$ / $2{\times}$
& Optimization frequency, track budget, solver iterations, and candidate-track multiplier for causal anchor-constrained BA. \\

Final anchored BA
& $10{,}000$ / $400$ / $3{\times}$
& Selected-track budget, solver iterations, and candidate-track multiplier for final submap refinement. \\

Preferred final-BA track observations
& $\geq 3$
& Prefer tracks observed in at least three images; two-view tracks may be retained as fallback. \\

Tail-anchor matching window
& 8
& Number of latest movable frames explicitly matched with each tail anchor before final BA. \\

Final-BA solver
&  Dense Schur
& Bundle-adjustment backend and linear solver used for final submap refinement. \\

Final-BA robust loss
& Soft-L1 ($1.5$ px)
& Robust loss and scale applied to final-BA reprojection residuals. \\

Final-BA depth filter
& $100.0$ m
&  Maximum-depth cutoff is applied during final BA. \\

\bottomrule
\end{tabular}
\end{table*}
\subsection{Algorithmic Details}
Algorithm~\ref{alg:anchor} summarizes the anchor localization stage used to produce {sparse anchor poses} for trajectory recovery. {The aerial visibility graph restricts expansion to views with overlapping aerial footprints, while the ground-first reconstruction schedule keeps the anchor group stable before aerial views are used for global metric alignment. The final BA fixes the reconstructed aerial cameras at their reference poses and refines the remaining local reconstruction, producing anchor poses directly in the aerial metric frame.}

\begin{algorithm}[t]
\caption{anchor localization}
\label{alg:anchor}
\small
\LinesNumbered
\KwIn{{Aerial SfM reconstruction $\mathcal{M}^a$, ground images $\{I_k^g\}$}}
\KwOut{{Anchor poses in the aerial metric frame}}

\tcp{{Select sparse anchor groups}}
$G_{\mathrm{vis}}^a \leftarrow$ {aerial visibility graph} of $\mathcal{M}^a$\;
$\{{A}_t\} \leftarrow$ {front/rear anchor groups} from the ground submaps\;

\ForEach{anchor group ${A}_t$}{
    \tcp{{Collect geometrically selected aerial support}}
    $\mathcal{R}_t \leftarrow \emptyset$\;
    
    \ForEach{ground image $I_k^g \in {A}_t$}{
        $\mathcal{C}_k \leftarrow$ retrieve aerial candidates for $I_k^g$\;
        {$I_{k,\mathrm{seed}}^a \leftarrow$ candidate with the largest RANSAC inlier count}\;
        {$\mathcal{R}_t \leftarrow \mathcal{R}_t \cup \{I_k^g,I_{k,\mathrm{seed}}^a\} \cup \operatorname{Nbr}(I_{k,\mathrm{seed}}^a)$}\;
    }

    \tcp{Recover anchor poses in the aerial metric frame}
    Reconstruct $\mathcal{R}_t$ with the {ground-first reconstruction schedule} and Aerial-MASt3R matches\;
    $\mathbf{S}_t \leftarrow \mathrm{Sim}(3)$ alignment from reconstructed aerial cameras to $\mathcal{M}^a$\;
    Transform the {local reconstruction} by $\mathbf{S}_t$\;
    {Refine the local reconstruction with final BA while keeping the aerial poses fixed}\;
}
\Return accepted anchor poses\;
\end{algorithm}

\subsection{Anchor Localization Ablations}
\subsubsection{Image Retrieval Model}

Image retrieval determines the aerial candidates passed to geometric verification and anchor localization. Table~\ref{tab:image_retrieval_model} compares NetVLAD, MegaLoc, SALAD, and AnyLoc under the same downstream pipeline. AnyLoc gives the most accurate anchor poses on both scenes; the other retrieval models can still register many anchor frames, but their poorer candidate neighborhoods lead to larger metric errors, especially on Ours-Campus and the HorizonGS-Road SALAD case.

\begin{table}[t]
\centering
\scriptsize
\setlength{\tabcolsep}{10pt}
\renewcommand{\arraystretch}{1.05}
\caption{Comparison of different image retrieval models in candidate aerial retrieval.}
{}
\label{tab:image_retrieval_model}
\resizebox{\linewidth}{!}{%
\begin{tabular}{@{}llcccc@{\hspace{8pt}}}
\toprule
Dataset & Method & ATE$\downarrow$ & RPE-t$\downarrow$ & RPE-r$\downarrow$ & Reg. \\
\midrule

\multirow{4}{*}{Ours-Campus}
& NetVLAD & \second{0.862} & 0.298 & 4.530 & 42/48 \\
& MegaLoc & 1.695 & 0.587 & 3.360 & \first{48/48} \\
& SALAD & 0.926 & \second{0.236} & \second{3.150} & \first{48/48} \\
& AnyLoc (ours used) & \first{0.058} & \first{0.021} & \first{0.330} & \first{48/48} \\
\midrule

\multirow{4}{*}{HorizonGS-Road}
& NetVLAD & \second{0.142} & \second{0.043} & \second{0.090} & \first{36/36} \\
& MegaLoc & 0.166 & 0.045 & 0.100 & \first{36/36} \\
& SALAD & 55.800 & 9.983 & 6.250 & \first{36/36} \\
& AnyLoc (ours used) & \first{0.098} & \first{0.033} & \first{0.080} & \first{36/36} \\

\bottomrule
\end{tabular}}
\end{table}

\begin{figure*}[t]
    \centering
    \includegraphics[width=1\linewidth]{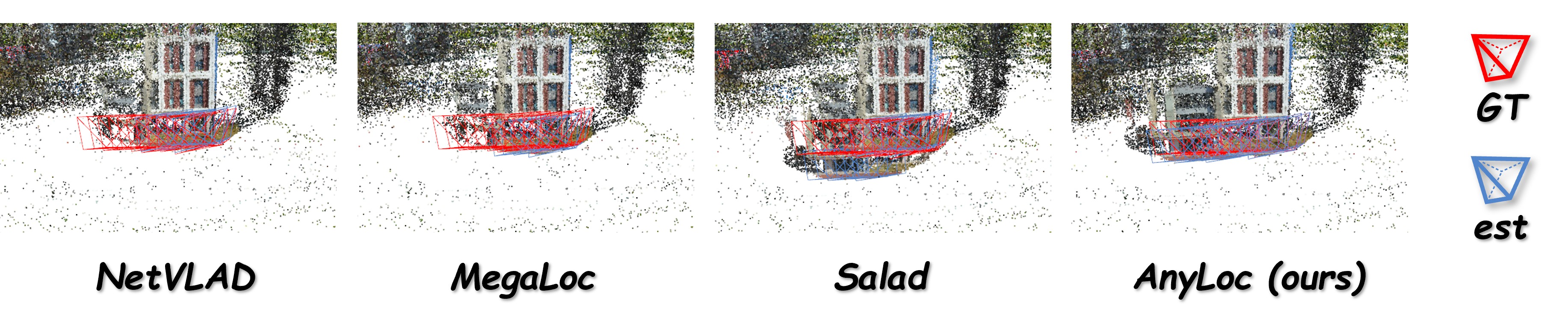}
    {}
    \caption{\textbf{Image retrieval ablation.} More reliable aerial candidate retrieval improves the geometric verification stage and reduces downstream anchor pose errors.}
    \label{fig:image_retrieval_model}
\end{figure*}

\subsubsection{Dense Matching Method}

Dense matching affects both the {ground-first reconstruction schedule} and aerial registration into the {local reconstruction}. Table~\ref{tab:dense_matching_method} shows that generic dense matchers often produce complete but metrically wrong reconstructions under the large aerial--ground viewpoint gap. MASt3R without aerial--ground adaptation fails to register anchor frames in these settings, while Aerial-MASt3R provides the cross-view correspondences needed for stable anchor localization.

\begin{table}[t]
\centering
\scriptsize
\setlength{\tabcolsep}{8pt}
\renewcommand{\arraystretch}{1.05}
\caption{Comparison of different dense matching methods for anchor localization.}
{}
\label{tab:dense_matching_method}
\resizebox{\linewidth}{!}{
\begin{tabular}{@{}llcccc@{\hspace{8pt}}}
\toprule
Dataset & Method & ATE$\downarrow$ & RPE-t$\downarrow$ & RPE-r$\downarrow$ & Reg. \\
\midrule

\multirow{4}{*}{Ours-Campus}
& Roma & 63.800 & 13.380 & 10.200 & \first{48/48} \\
& RomaV2 & \second{34.860} & \second{2.553} & \second{0.590} & \first{48/48} \\
& MASt3R & -- & -- & -- & 0/48 \\
& Aerial-MASt3R (ours used) & \first{0.058} & \first{0.021} & \first{0.330} & \first{48/48} \\
\midrule

\multirow{4}{*}{HorizonGS-Road}
& Roma & \second{0.159} & \second{0.051} & \second{0.160} & \first{36/36} \\
& RomaV2 & 20.020 & 5.819 & 3.730 & \first{36/36} \\
& MASt3R & -- & -- & -- & 0/36 \\
& Aerial-MASt3R (ours used) & \first{0.098} & \first{0.033} & \first{0.080} & \first{36/36} \\

\bottomrule
\end{tabular}}
\end{table}

\begin{figure*}[t]
    \centering
    \includegraphics[width=1\linewidth]{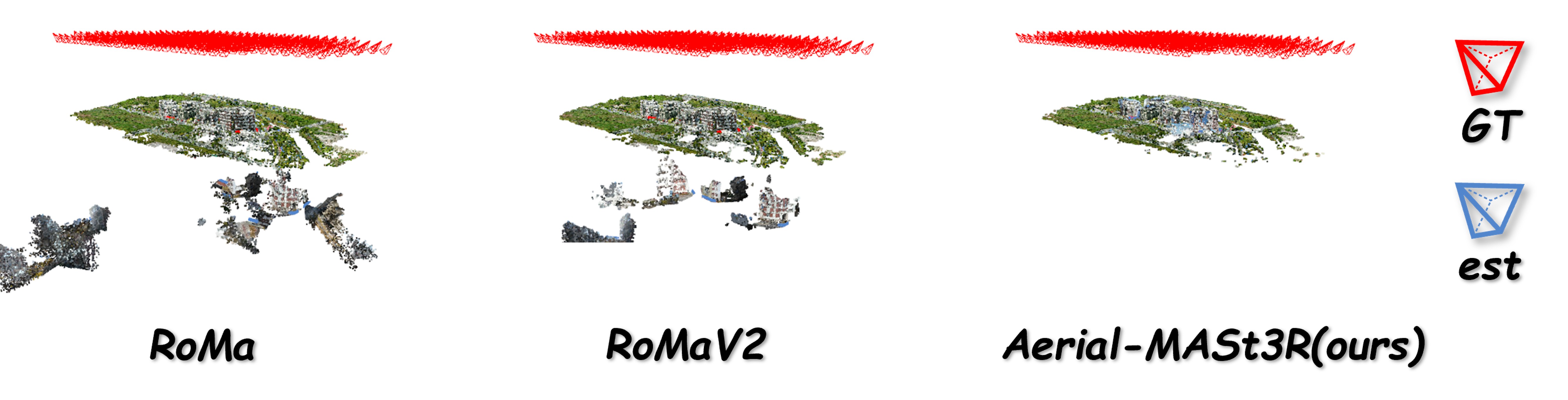}
    {}
    \caption{\textbf{Dense matching ablation.} Aerial-MASt3R produces more reliable cross-view matches than general-purpose dense matchers for anchor localization.}
    \label{fig:dense_matching_method}
\end{figure*}

\subsubsection{Aerial--Ground Image Composition}

Table~\ref{tab:aerial_ground_setting} isolates whether directly feeding aerial and ground images to a feed-forward reconstruction model is sufficient for anchor localization. VGGT reconstructs short ground-only groups accurately, but its pose estimates degrade sharply when aerial images are added because the mixed set introduces severe scale and viewpoint changes. SkyAnchor instead keeps each anchor group stable and uses verified aerial views as metric support, which yields low errors across all three anchor groups.

\begin{table*}[t]
\centering
\scriptsize
\setlength{\tabcolsep}{10pt}
\renewcommand{\arraystretch}{1.0}
\caption{\textbf{Comparison of pose estimation accuracy under different aerial--ground image settings. }Each group contains 6 ground images and a different number of aerial images.}
{}
\label{tab:aerial_ground_setting}
\resizebox{\linewidth}{!}{
\begin{tabular}{@{}llccccccccc@{}}
\toprule
\multirow{3}{*}{Method} 
& \multirow{3}{*}{Input Images} 
& \multicolumn{3}{c}{Anchor Group 1} 
& \multicolumn{3}{c}{Anchor Group 2} 
& \multicolumn{3}{c}{Anchor Group 3} \\
\cmidrule(lr){3-5} \cmidrule(lr){6-8} \cmidrule(l){9-11}
& 
& \multicolumn{3}{c}{6 Ground + 17 Aerial}
& \multicolumn{3}{c}{6 Ground + 4 Aerial}
& \multicolumn{3}{c}{6 Ground + 8 Aerial} \\
\cmidrule(lr){3-5} \cmidrule(lr){6-8} \cmidrule(l){9-11}
& 
& ATE$\downarrow$ & RPE-t$\downarrow$ & RPE-r$\downarrow$
& ATE$\downarrow$ & RPE-t$\downarrow$ & RPE-r$\downarrow$
& ATE$\downarrow$ & RPE-t$\downarrow$ & RPE-r$\downarrow$ \\
\midrule
VGGT 
& Ground only
& 0.003 & 0.003 & 0.240
& 0.011 & 0.010 & 0.200
& 0.006 & 0.007 & 0.190 \\
VGGT 
& Aerial + Ground
& 10.34 & 3.429 & 8.160
& 19.71 & 6.109 & 12.22
& 14.73 & 10.01 & 9.210 \\
Ours 
& Aerial + Ground
& 0.030 & 0.020 & 0.080
& 0.025 & 0.033 & 0.190
& 0.034 & 0.025 & 0.120 \\
\bottomrule
\end{tabular}}
\end{table*}

\subsection{{Anchor Localization Runtime}}
Table~\ref{tab:anchor_reconstruction_runtime} reports the runtime of different methods for anchor localization on various datasets. The runtime differences mainly arise from two factors. First, COLMAP and InstantSfM employ GPU-accelerated SIFT-based sparse feature extraction, which is generally faster than the MASt3R-based dense matching strategies used by MASt3R-SfM, MP-SfM, and our method. Second, due to the challenging nature of cross-view localization, several methods fail to complete the reconstruction on some datasets or reconstruct only a subset of the images, resulting in shorter runtimes than our method in those cases. In contrast, our method consistently completes the reconstruction across all datasets. As shown in the table, while ensuring reconstruction accuracy and stability, our method still maintains a practical and reasonable runtime.

\begin{table*}[!t]
\footnotesize
    \setlength\tabcolsep{10pt}
    \centering
    \renewcommand\arraystretch{1.}
    \caption{\textbf{Runtime comparison of anchor localization.} We report the runtime of each method on seven datasets.}
    {}
    \label{tab:anchor_reconstruction_runtime}
    \resizebox{\linewidth}{!}{
\begin{tabular}{l *{7}{c}}
\toprule
Method
& Ours-Campus& HorizonGS-Park& HorizonGS-Road& ISPRS-Rathaus& ISPRS-Stadthaus& ISPRS-Zeche& ALoG-Uni10k \\
\midrule
{COLMAP}& -& -& -& 15m 38.79s& 26m 13.78s& 3m 1.59s& 21m 2.48s \\
{InstantSfM}& -& 12m 32.41s& -& 14m 59.71s& 6m 23.12s& 1m 51.72s& 9m 50.14s \\
{MASt3R-SfM}& 26m 31.74s& 42m 52.97s& 31m 50.81s& 25m 42.85s& 17m 51.23s& 7m 10.41s& 24m 26.19s \\
{MP-SfM}& -& -& -& 19h 38m 16s& -& -& 23h 14m 24s \\
\midrule
Ours& 48m 45.18s& 1h 20m 44.63s& 37m 16.90s& 42m 51.72s& 1h 6m 8.92s& 15m 34.99s& 1h 3m 24.88s \\
\bottomrule
\end{tabular}}
\end{table*}

\begin{table}[!ht]
\centering
\scriptsize
\setlength{\tabcolsep}{5pt}
\renewcommand{\arraystretch}{1.0}
\caption{Runtime breakdown for one ALoG-Uni10k ground submap. The submap contains 100 input frames and 47 selected keyframes. All times are wall-clock seconds. Online averages are normalized by the 100 input frames, while the {anchor-constrained BA} is a global refinement and is not reported as a per-frame average. Indented entries are diagnostic subcomponents and are not additive.}
\label{tab:runtime_breakdown}
\resizebox{\linewidth}{!}{
\begin{tabular}{lccc}
\toprule
Stage / component & Time (s) & Share & Avg. / input frame \\
\midrule
Total submap & 280.24 & 100.0\% & -- \\

\rowcolor{gray!15}
\multicolumn{4}{l}{\textbf{Online stages}} \\
Online subtotal & 152.54 & 54.4\% & 1.53 \\
Seeded bootstrap & 25.26 & 9.0\% & 0.25 \\
Incremental registration & 78.18 & 27.9\% & 0.78 \\
\quad -Feature extraction & 10.45 & 3.7\% & 0.10 \\
\quad -Keyframe decision matching & 34.50 & 12.3\% & 0.35 \\
\quad -3D-aware keyframe selection & 0.49 & 0.2\% & 0.005 \\
\quad -PnP-RANSAC & 0.57 & 0.2\% & 0.006 \\
\quad -One-frame mini-BA & 4.92 & 1.8\% & 0.05 \\

\midrule
\rowcolor{gray!15}
\multicolumn{4}{l}{\textbf{Global refinement}} \\
{Anchor-constrained BA} & 127.70 & 45.6\% & -- \\
\quad  -solver & 2.55 & 0.9\% & -- \\
\bottomrule
\end{tabular}}
\end{table}

\begin{table}[t]
\centering
\scriptsize
\setlength{\tabcolsep}{10pt}
\renewcommand{\arraystretch}{0.9}
\caption{Additional ablation on anchor group size.}
{}
\label{tab:ablation_traj_anchor}
\resizebox{\linewidth}{!}{
\begin{tabular}{@{}llccc@{\hspace{14pt}}}
\toprule
 & Anchor size & ATE$\downarrow$ & RPE-t$\downarrow$ & RPE-r$\downarrow$ \\
\midrule
\multirow{3}{*}{Trajectory } 
&(2, 2) & 0.325 & 0.832 & 0.461 \\
&(4, 4) & 0.184 & 0.559 & 0.351 \\
&(6, 6) & 0.176 & 0.510 & 0.306 \\

\bottomrule
\end{tabular}}
\end{table}

\section{Anchor-Constrained Trajectory Recovery Details}
\label{app:trajectory_details}

\subsection{Algorithmic Details}
Algorithm~\ref{alg:anchor_trajectory} gives the implementation details of the trajectory recovery stage. Compared with a standard incremental PnP pipeline, three details are important. First, every submap is initialized by a seeded bootstrap in which the front anchor poses are fixed, so the local map starts directly in the aerial metric frame. Second, each incoming keyframe is registered against reference keyframes selected by {3D-aware keyframe selection} rather than by temporal proximity alone. Third, all anchor frames are restored to their localized anchor poses before the final {anchor-constrained BA}, where they act as fixed boundary conditions for optimizing the non-anchor cameras.

\begin{algorithm}[h]
\caption{Anchor-Constrained Trajectory Recovery}
\label{alg:anchor_trajectory}
\scriptsize
\LinesNumbered
\KwIn{Ground-view sequence $\{I_i^g\}$, {anchor poses $\{\mathbf{T}_{i,\mathrm{anc}}^g\}$, submaps $\{\mathcal{S}_m\}$}}
\KwOut{Ground trajectory in the aerial metric frame}

$\mathcal{K} \leftarrow \emptyset$\;
\ForEach{submap $\mathcal{S}_m$}{
    \tcp{Seed the submap directly in the aerial metric frame}
    {Initialize local ground map $\mathcal{M}_m^g$}\;
    {$\mathcal{B}_m^f,\mathcal{B}_m^r \leftarrow$ front and rear anchor frames in $\mathcal{S}_m$; $\mathcal{B}_m\leftarrow\mathcal{B}_m^f\cup\mathcal{B}_m^r$}\;
    {$\mathcal{Q}_m \leftarrow$ first $N_{\mathrm{boot}}$ keyframes of $\mathcal{S}_m$, including $\mathcal{B}_m^f$}\;
    Match all pairs in $\mathcal{Q}_m$ and build local feature tracks\;
    Run seeded mini-BA on $\mathcal{Q}_m$ with $\mathcal{B}_m^f$ fixed\;
    {Insert the bootstrap keyframes into $\mathcal{M}_m^g$ and triangulate local 3D points}\;

    \ForEach{remaining frame $I_i^g \in \mathcal{S}_m$}{
        \tcp{Register selected frames with {3D-aware keyframe selection}}
        {Extract dense features for $I_i^g$}\;
        \If{{$I_i^g$ passes the keyframe test or is an anchor/test frame}}{
            {$\mathcal{K}_i^{\mathrm{ref}} \leftarrow$ reference keyframes selected from a wider previous-keyframe window by 3D-point-supported matches}\;
            {Form 2D--3D pairs $(\mathbf{X}_j,\mathbf{u}_{ij})$ from matches between $I_i^g$ and $\mathcal{K}_i^{\mathrm{ref}}$}\;
            {Optionally filter these pairs by reference depth and match confidence}\;
            {$\hat{\mathbf{T}}_i^g \leftarrow$ PnP-RANSAC on the retained pairs, followed by one-frame mini-BA}\;
            \If{$I_i^g$ is an anchor frame}{
                {$\hat{\mathbf{T}}_i^g \leftarrow \mathbf{T}_{i,\mathrm{anc}}^g$}\;
            }
            \If{{$\hat{\mathbf{T}}_i^g$ is accepted}}{
                {Insert $(I_i^g,\hat{\mathbf{T}}_i^g)$ into $\mathcal{M}_m^g$ and update local 3D points}\;
            }
        }
    }

    \tcp{Close the submap with {anchor-constrained BA}}
    {Reset every anchor frame in $\mathcal{M}_m^g$ to $\mathbf{T}_{i,\mathrm{anc}}^g$}\;
    {Generate candidate tracks from non-anchor and fixed anchor frames}\;
    {Select long, valid, depth-bounded tracks, prioritizing anchor-connected observations}\;
    {$\mathcal{O}_m \leftarrow$ frame--point observations induced by the selected tracks}\;
    Optimize non-anchor poses and 3D points with all anchor poses fixed\;
    {$\mathcal{K} \leftarrow \mathcal{K} \cup$ optimized keyframes from $\mathcal{M}_m^g$}\;
}
\KwRet{$\mathcal{K}$}\;
\end{algorithm}

During incremental registration, 3D-aware keyframe selection is a lightweight pre-filter before the more expensive matcher and PnP-RANSAC. The method checks a wider window of previous keyframes and scores each candidate by the number of cheap descriptor matches whose reference keypoints already have triangulated 3D points. This avoids selecting recent frames that share many 2D features but provide too few usable 2D--3D correspondences, which is a common failure mode in open scenes with low parallax.

The final {anchor-constrained BA} is designed to make the {rear anchor poses} active constraints rather than only diagnostic poses. Before optimization, all anchor frames are reset to their localized anchor poses and placed in the fixed-camera block, while the remaining cameras are optimized together with local 3D points. Candidate tracks are first over-generated and then selected to favor multi-view observations, especially tracks touching both anchor and non-anchor frames. This transfers the aerial metric frame from the {front and rear anchor groups} into the interior of each submap before the optimized submaps are concatenated.

\subsection{Trajectory Recovery Runtime}
Table~\ref{tab:runtime_breakdown} reports the runtime breakdown on one representative ALoG-Uni10k ground submap. The online trajectory recovery stages take 152.54 seconds in total, corresponding to 1.53 seconds per input frame. Within this online part, incremental registration is the dominant cost, taking 78.18 seconds, while the seeded bootstrap takes 25.26 seconds and is executed only once per submap. The 3D-aware keyframe selection and PnP-RANSAC steps are lightweight, taking only 0.49 seconds and 0.57 seconds respectively over the whole submap. This indicates that the additional geometric checks used by our robust registration introduce negligible overhead compared with feature matching and incremental pose refinement.

The {anchor-constrained BA} takes 127.70 seconds and accounts for 45.6\% of the total submap runtime. Unlike the online stages, this is a global submap refinement performed once after all frames have been registered, so we do not report a per-frame average for it. Its cost reflects the construction and optimization of {anchor-connected multi-view tracks}, which improves metric consistency before the optimized submap is passed to scene update.

\subsection{Anchor Group Size}
Table~\ref{tab:ablation_traj_anchor} studies how many localized anchor frames are needed at each submap boundary. Using only two front and two rear anchor frames gives noticeably weaker trajectory constraints, since each anchor group has limited multi-view geometry. Increasing the group size to $(4,4)$ substantially reduces all errors, indicating that consecutive anchor frames provide a more stable local reference for both seeded bootstrap and {anchor-constrained BA}. The $(6,6)$ setting gives the best result, with smaller additional gains over $(4,4)$, suggesting that anchor constraints become saturated once each boundary contains enough local overlap and parallax. We therefore use six front and six rear anchor frames as the default setting in our experiments.

\section{Additional Results}
This section provides additional qualitative results. Fig.~\ref{fig:bapoint} visualizes the dense BA points used for {ground Gaussian initialization and scene update}. Fig.~\ref{fig:addianchor} shows additional anchor localization comparisons, Fig.~\ref{fig:additaj} presents more trajectory recovery comparisons, and Fig.~\ref{fig:addivis} provides additional visualizations of our {scene update} results.

\begin{figure}[t]
    \centering
    \includegraphics[width=1\linewidth]{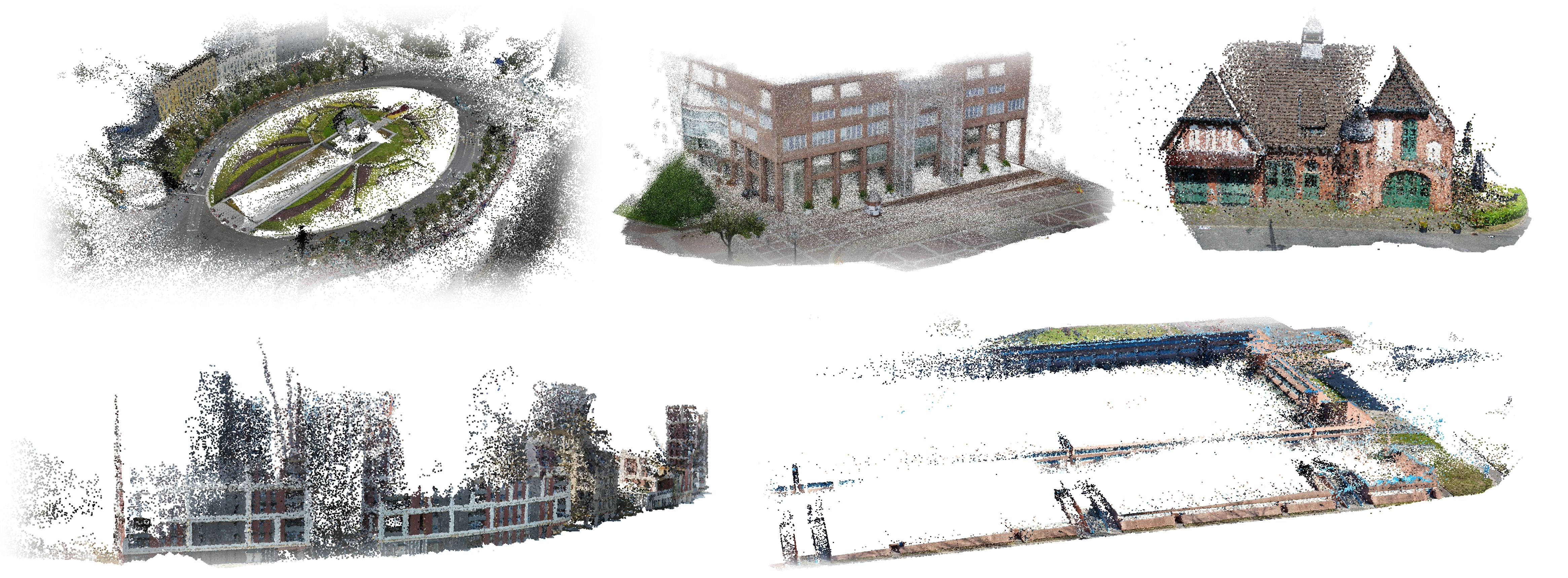}
    \caption{Visualization of our dense BA points for {ground Gaussian initialization and scene update}.}
    \label{fig:bapoint}
\end{figure}

\newpage
\begin{figure*}[t]
    \centering
    \includegraphics[width=0.9\linewidth]{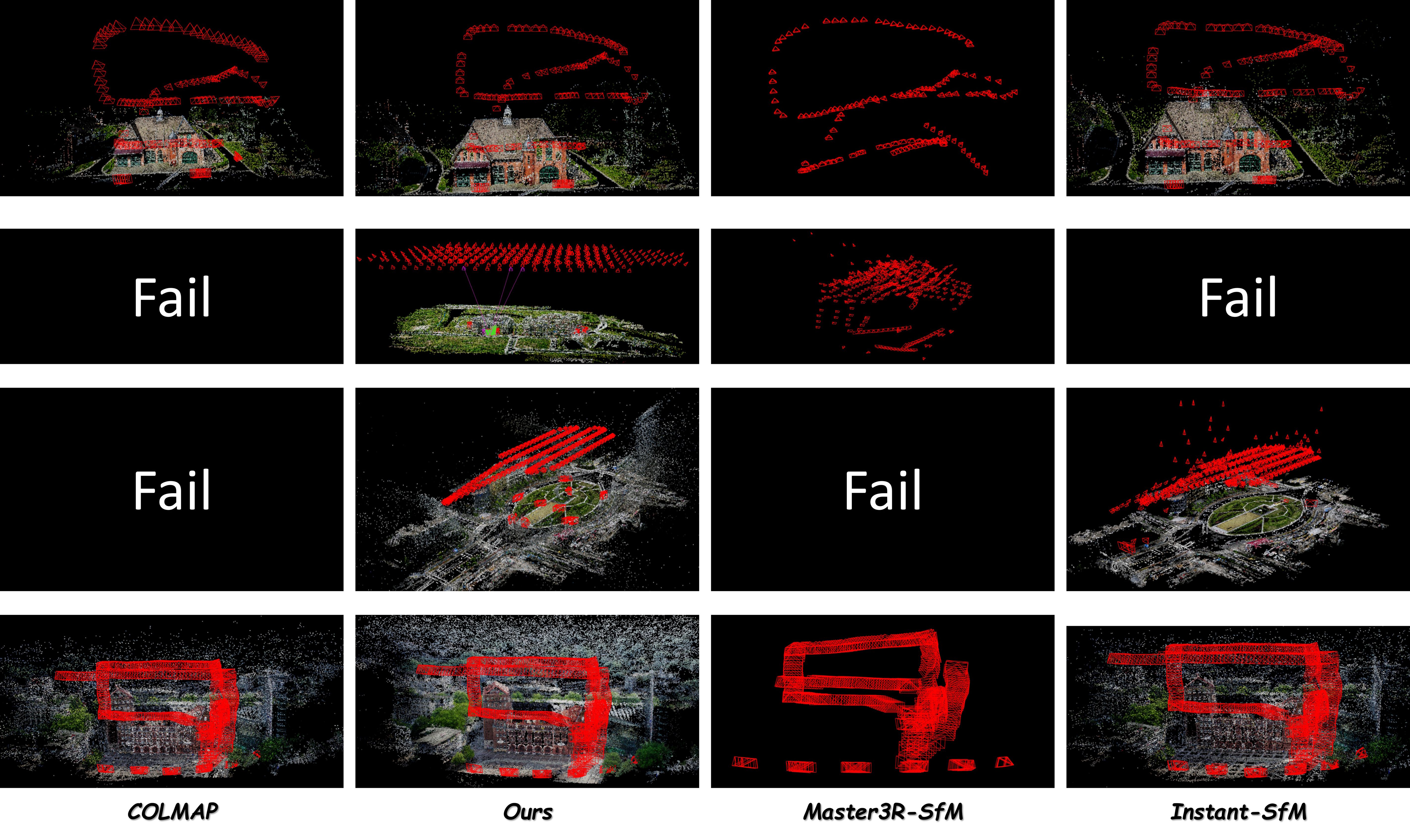}
    \caption{Additional comparison of anchor localization.}
    \label{fig:addianchor}
\end{figure*}

\begin{figure*}[t]
    \centering
\includegraphics[width=0.9\linewidth]{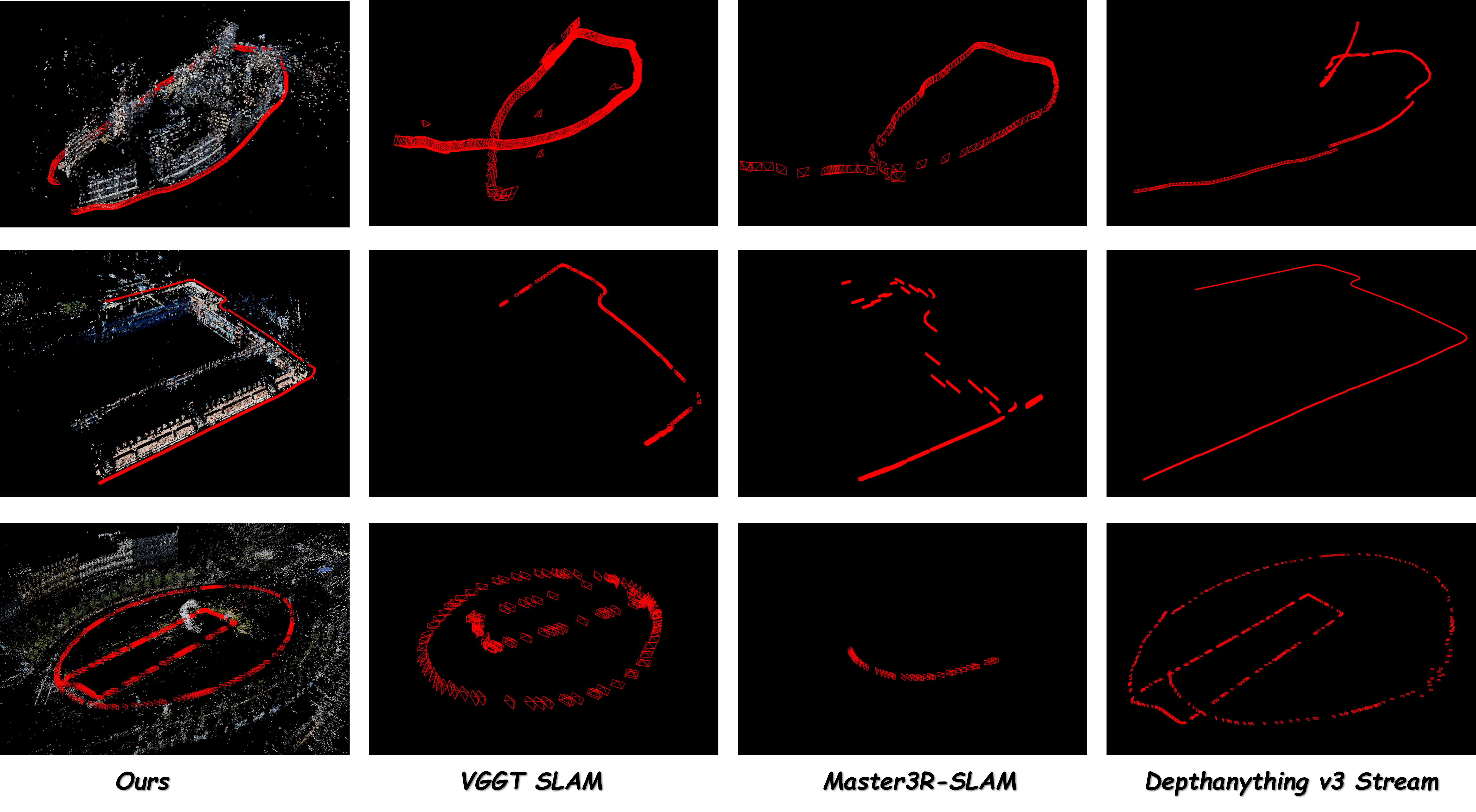}
    \caption{Additional comparison of {ground trajectories}.}
    \label{fig:additaj}
\end{figure*} 

\begin{figure*}[t]
    \centering
\includegraphics[width=1\linewidth]{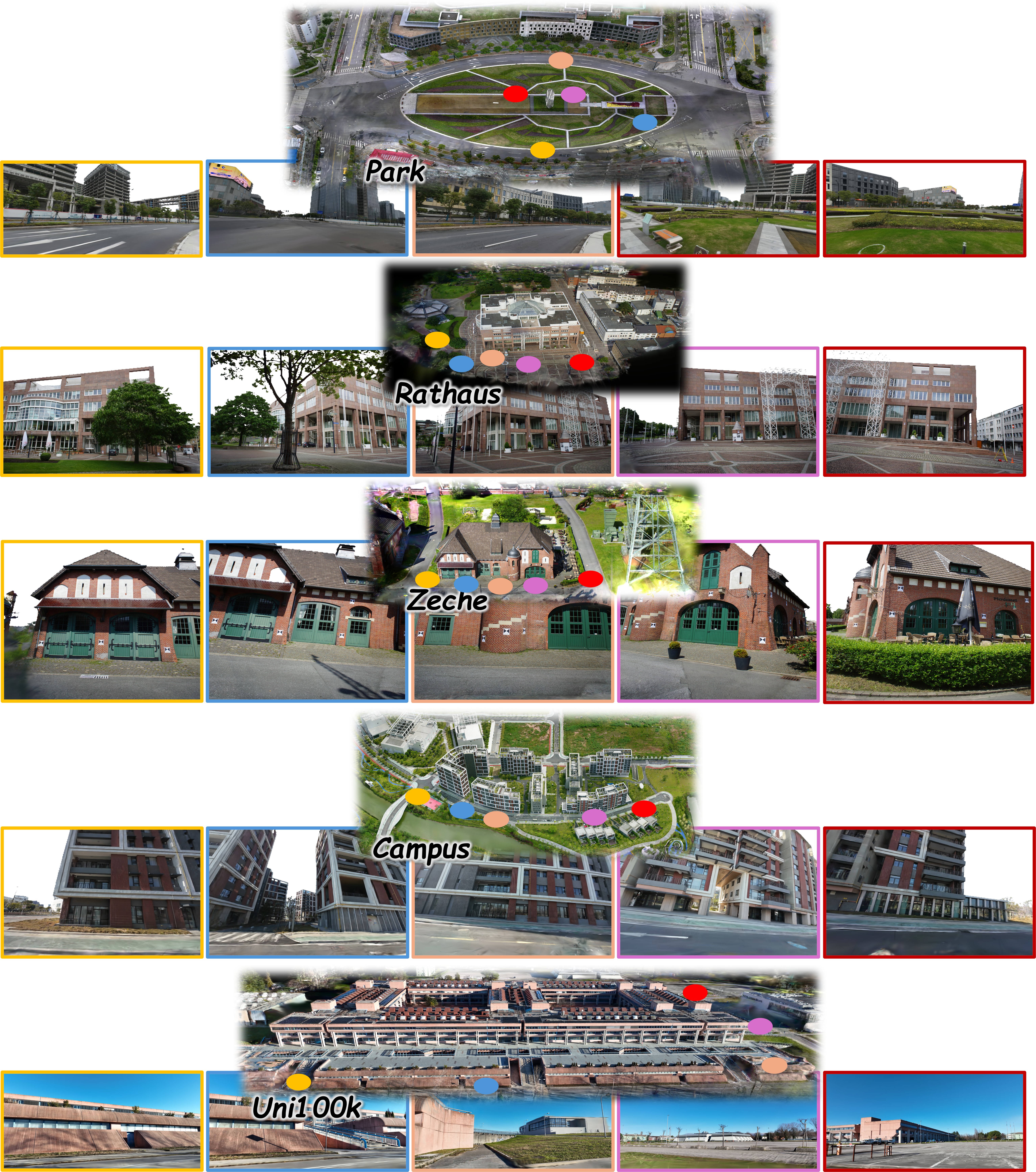}
    \caption{Additional visualization of our {scaffold-preserving scene update results}.}
    \label{fig:addivis}
\end{figure*}

\clearpage

\end{document}